\documentclass[10pt,twocolumn,letterpaper]{article}
\pdfoutput=1
\usepackage[pagenumbers]{iccv} % To force page numbers, e.g. for an arXiv version
\usepackage{amsmath}
\usepackage{float}
\usepackage{colortbl}
\usepackage{xcolor}
\definecolor{iccvblue}{rgb}{0.21,0.49,0.74}
\usepackage[pagebackref,breaklinks,colorlinks,allcolors=iccvblue]{hyperref}
\usepackage{pifont}

\def\paperID{2752} % *** Enter the Paper ID here
\def\confName{ICCV}
\def\confYear{2025}

\title{LightHeadEd: Relightable \& Editable Head Avatars from a Smartphone}

\author{Pranav Manu*\\
IIIT Hyderabad\\
{\tt\small pranav.m@research.iiit.ac.in}
\and
Astitva Srivastava\\
IIIT Hyderabad\\
{\tt\small astitva.s@research.iiit.ac.in}
\and
Amit Raj\\
Google Research\\
{\tt\small amitrajs@google.com}
\and
Varun Jampani\\
Stability AI\\
{\tt\small varunjampani@gmail.com}
\and
Avinash Sharma\\
IIT Jodhpur\\
{\tt\small avinashsharma@iitj.ac.in}
\and
P.J. Narayanan\\
IIIT Hyderabad\\
{\tt\small pjn@iiit.ac.in}
}

\begin{document}
% \maketitle
\twocolumn[{%
\renewcommand\twocolumn[1][]{#1}%
\maketitle
\begin{center}
    \centering
    \captionsetup{type=figure}
    \includegraphics[width=\linewidth]{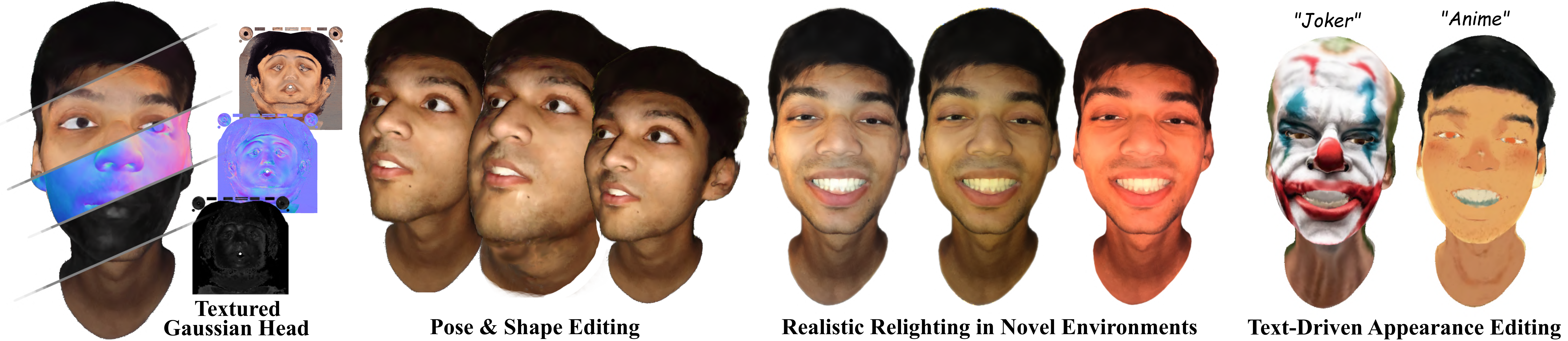}
    \captionof{figure}{We introduce LightHeadEd to reconstruct realistic head avatars with editing \& relighting support.}
    \label{fig:teaser}
\end{center}%
}]
\begin{abstract}
Creating photorealistic, animatable, and relightable 3D head avatars traditionally requires expensive Lightstage with multiple calibrated cameras, making it inaccessible for widespread adoption. To bridge this gap, we present a novel, cost-effective approach for creating high-quality relightable head avatars using only a smartphone equipped with polaroid filters. Our approach involves simultaneously capturing cross-polarized and parallel-polarized video streams in a dark room with a single point-light source, separating the skin's diffuse and specular components during dynamic facial performances. We introduce a hybrid representation that embeds 2D Gaussians in the UV space of a parametric head model, facilitating efficient real-time rendering while preserving high-fidelity geometric details. Our learning-based neural analysis-by-synthesis pipeline decouples pose and expression-dependent geometrical offsets from appearance, decomposing the surface into albedo, normal, and specular UV texture maps, along with the environment maps. We collect a unique dataset of various subjects performing diverse facial expressions and head movements. 
% We aim to bring scalability to polarized facial performance capture, while significantly reducing the complexity \& cost of the setup.
% , democratizing high-quality relightable head avatar creation.
\end{abstract}    
\section{Introduction}
\label{sec:intro}

% \textcolor{cyan}{[Problem statement \& its importance]} 
Relightable, animatable, and editable 3D human head avatars have gained significant popularity in recent years, serving as a versatile and immersive communication medium for the Metaverse, mixed reality platforms, telepresence and beyond.  However, creating personalized head avatars with relighting capabilities is challenging due to their dependence on exhaustive multiview data, typically captured using volumetric light-stage systems. These capture systems, being highly sophisticated, expensive, and compute-intensive, are often limited to high-budget production studios. In contrast, enabling affordable 3D head avatars using just a single commodity smartphone could democratize their access to end users, unlocking a wide array of applications in communication and entertainment.

% Such capture systems are highly sophisticated, expensive \& compute-intensive, restricting their access to high-budget production studios. The ability to create affordable realistic, animatable \& relightable head avatars using a single commodity smartphone would democratize their access, opening up a wide range of applications in communication and entertainment for the end user.\\
% 
% 
% % These realistic avatars enable natural communication through facial expressions and non-verbal cues, while being bandwidth-efficient \cite{Ma2021PixelCA}. 
% The first challenge in creating high-quality relightable avatars involves the dynamic acquisition of dynamically evolving 3D facial geometry and appearance, traditionally using carefully calibrated volumetric light-stage capture systems \cite{10.1145/3355089.3356571}, aiming to disentangle the geometry and appearance of the face. Given exhaustive visual data from the polarized cameras, differential physically-based rendering is used to reconstruct the facial surface details and estimate material parameters in an analysis-by-synthesis manner.
% 
% 
% \textcolor{cyan}{[Key existing works and their drawbacks which our method addresses]} 
A handful of existing methods \cite{10.1145/3528223.3530143, li2024uravatar} attempt to create personalized 3D head avatars from a monocular RGB video captured using a smartphone/webcam. However, they primarily depend on a universal prior model, trained on an exhaustive light-stage dataset of multiple subjects to estimate disentangled facial geometry \& reflectance. Such heavy dependence on a dataset/model significantly undermines the affordability and hinders the generalizability to unseen demographics \& appearances, while still demanding expensive preprocessing and test-time fine-tuning of the model. To eliminate reliance on any prior, several other approaches \cite{baert2024spark, Zheng2023pointavatar, Gafni_2021_CVPR} employ inverse physically-based rendering (PBR) to reconstruct a personalized head avatar in a self-supervised analysis-by-synthesis manner. Specifically, \cite{baert2024spark} uses a parametric head model, FLAME \cite{FLAME:SiggraphAsia2017}, to reconstruct a relightable parametric head mesh from several unconstrained RGB videos of a person. 
Though FLAME offers animation support via manipulation of low-dimensional pose/expression parameters, it provides only a coarse approximation of the facial geometry and does not model deviations/offsets from the underlying surface (skin deformations, hair, beard, etc.).  In order to learn these offsets, recent methods such as \cite{ma2024gaussianblendshapes, chen2023monogaussianavatar} propose to combine 3DGS~\cite{kerbl3Dgaussians} and FLAME head mesh. They propose to sample 3D Gaussian-based splats over a FLAME mesh to learn person-specific details while allowing control over expression/pose. However, these methods are resource intensive, lack relighting support and produce inaccurate geometrical details. In terms of appearance editing, none of the existing monocular methods allow control over the appearance manipulation.
% \vjnote{Last two paragraphs on related works is a bit lengthy for the intro. We can condense a bit if space is needed.}
% 
% 
% To recover high-fidelity geometry detailed enough to model expression-dependent wrinkles, existing learning-based non-parametric approaches \cite{Zheng2023pointavatar, Gafni_2021_CVPR} learn offsets over the FLAME head mesh, either using point-cloud, implicit representations (e.g. SDF) or neural representations (e.g. NeRF). However, these representations are not efficient enough for real-time applications and are difficult to integrate with standard graphics pipelines. Recently proposed 3DGS\cite{kerbl3Dgaussians} enables real-time novel-view rendering of  3D scenes and has been successfully applied to head avatars in a multi-view setting \cite{qian2023gaussianavatars, saito2024rgca}.
% 
% Unlike existing approaches that estimate parametric reflectance parameters via inverse rendering, our approach directly models learnable radiance transfer that incorporates global light transport in an efficient manner for real-time rendering. However, learning such a complex light transport that can generalize across identities is non-trivial. A phone scan in a single environment lacks sufficient information to infer how the head would appear in general environments.
% 
% 
\begin{figure}
    \centering
    \includegraphics[width=\linewidth]{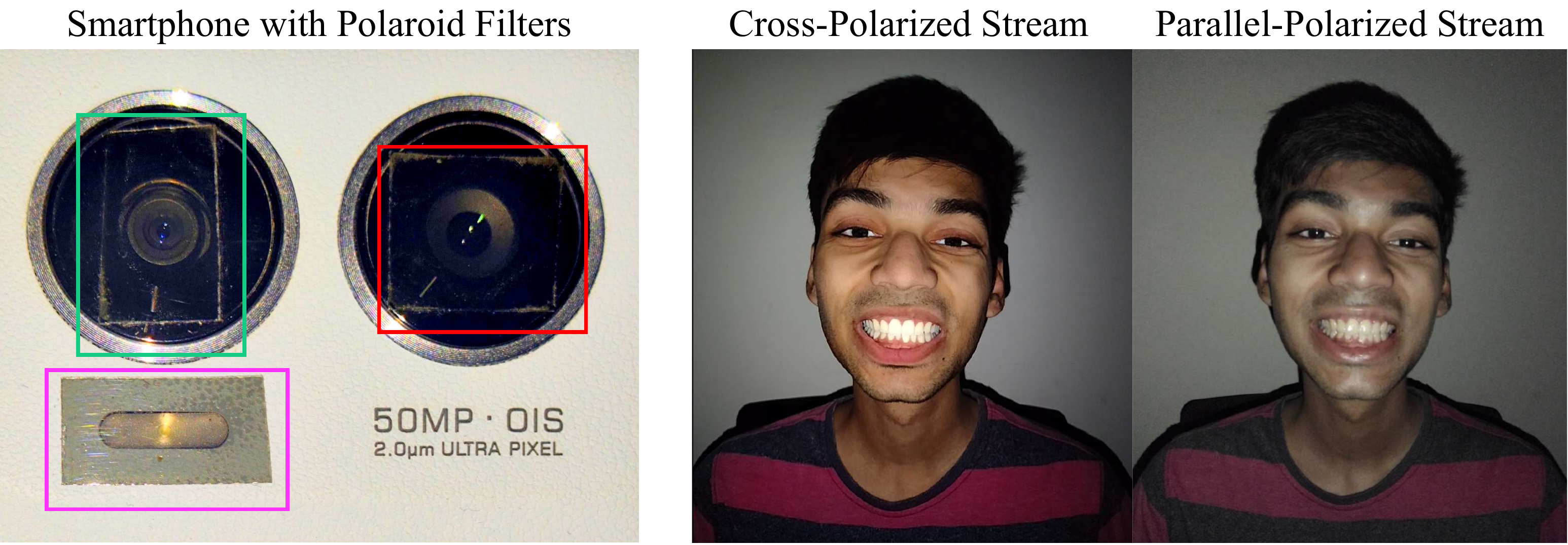}
    \caption{Proposed dynamic capture setup: Smartphone equipped with polaroid filters (left); Cross-Polarized \& Parallel-Polarized Monocular Video Streams (right).
    }
    \label{fig:setup}
\end{figure}

In this paper, we present a novel approach to effortlessly create animatable, editable and relightable head avatars from a commodity smartphone. Instead of relying on a well-calibrated light-stage for the dynamic acquisition of facial performances, we propose a cost-effective and calibration-free capture process to gather realistic head deformations and facial reflectance information. In order to enable relighting, animation and editing support, we propose a novel textured Gaussian head avatar representation, along with an effective self-supervised training methodology to learn a personalized head avatar with high-quality geometry and decomposed albedo, normal and roughness UV maps.
Our motivation stems from the fact that the key feature of a light-stage capture setup is multiple RGB cameras equipped with polaroid filters, aiming to decompose skin’s diffuse and specular response to the illumination. Leveraging the same principle, we propose to equip the dual cameras and flashlight of a smartphone with an inexpensive polaroid film, as shown in \autoref{fig:setup}, resulting in a scalable capture process to acquire realistic head deformations of a subject with decomposed surface illumination, in the form of monocular videos.
 % 
  % we use two back cameras of a smartphone mounted on a tripod to simultaneously capture one cross-polarized and one parallel-polarized monocular video stream of the human subject enacting diverse head movements and facial expressions, while treating the flashlight as a single point-light source in a dark room. Additionally, unlike \cite{azinovic2022polface}, we delegate the requirement of a color-checker and light fall-off estimation by carefully designing the self-supervised training methodology
% 

% After dynamic polarized data acquisition, the next goal is to reconstruct a personalized relightable head avatar.
Next, we aim to use the captured polarization data to reconstruct the relightable 3D head avatar of the subject with plausible temporal consistency. 
We first track a FLAME head mesh over the monocular videos, and propose to use 2DGS~\cite{Huang2DGS2024} to model details like skin, hair, beard, etc. Unlike \cite{ma2024gaussianblendshapes, chen2023monogaussianavatar} which combine 3DGS with FLAME head mesh, our 2DGS-based representation supports direct surface regularization through normal constraints, achieving high-fidelity facial details while enabling real-time rendering. We embed the 2D Gaussian disks in FLAME's UV space to learn the Gaussian attributes in the form of UV maps for easy animation and texturing support. Learning appearance in UV space helps eliminate the need for memory-intensive SH-coefficients and more importantly, enables flexible appearance editing by altering the albedo map. We decouple reflectance in the form of albedo, normals and roughness UV maps. Additionally, to handle deformations and appearance changes, we propose to learn expression-dependent residual UV maps. Furthermore, we also learn an environment cubemap to support relighting in arbitrary lighting conditions. 
% 
% \vjnote{The previous text is more like a method overview and best suited at the beginning of the methods rather than the intro. Intro should highlight the key innovations and insights.}
% We also employ bilateral grid to rectify color disparity between the cross and parellel-polarized frames. 
% Unlike two-stage optimization scheme proposed by \cite{azinovic2022polface}, we learn reflectance parameters via a single self-supervised joint optimization process, achieving faster training.
% Though \cite{ma2024gaussianblendshapes} offers comparable training speed, our decomposed $uv$-based representation is far more storage efficient than \cite{ma2024gaussianblendshapes}. 

The proposed method for head avatar capture \& reconstruction is highly efficient and enables affordable relightable heads with several intriguing features. We demonstrate superior quality results, backed by comprehensive quantitative and ablative analysis, while showcasing several useful applications, such as shape editing \& text-guided appearance editing.
\noindent
In summary, our contributions include: % are as follows:
\begin{itemize}
    \item We propose an affordable and scalable process for the dynamic acquisition of polarized facial performances from a commodity smartphone equipped with a polaroid film.
    \item We introduce a novel 2DGS-based head avatar representation with relighting, texture mapping \& editing support; and a novel methodology to learn the aforementioned representation from a polarized monocular video sequence.
    \item We collect a first-of-its-kind dataset of polarized facial performances with diverse facial expressions and head movements. 
    
    % \vjnote{If the dataset is not big, this claim as key contribution may backfire.}
    % We will release the code and dataset to promote research in these areas.
\end{itemize}

\noindent
We plan to release the code and dataset publicly to drive the advancements in relightable head avatars research.

% \todo{PJN: Intro is too detailed and contains a good review of the most related prior work. As Varun points out, it also gives a longer-than-usual intro to the method. While these make the intro very readable, it will take up too much space. 
% }

\begin{figure*}
    \centering
    \hspace*{0.05cm}
    \includegraphics[width=\linewidth]{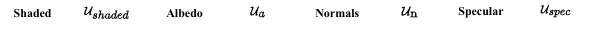}
    \includegraphics[width=\linewidth]{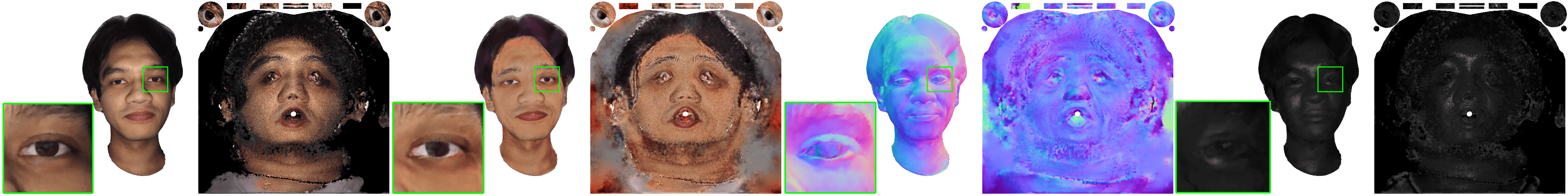}
    \includegraphics[width=\linewidth]{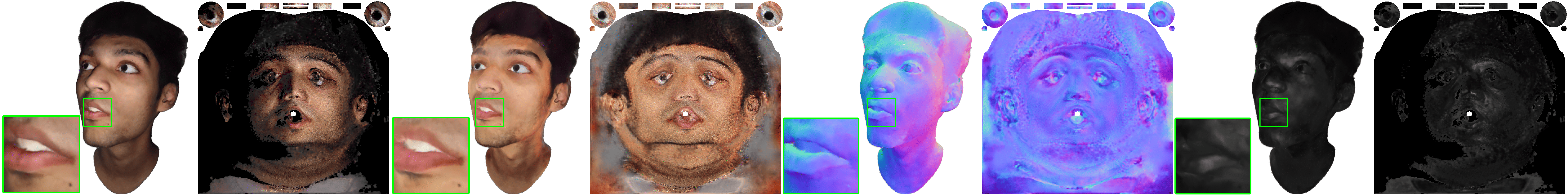}
    \includegraphics[width=\linewidth]{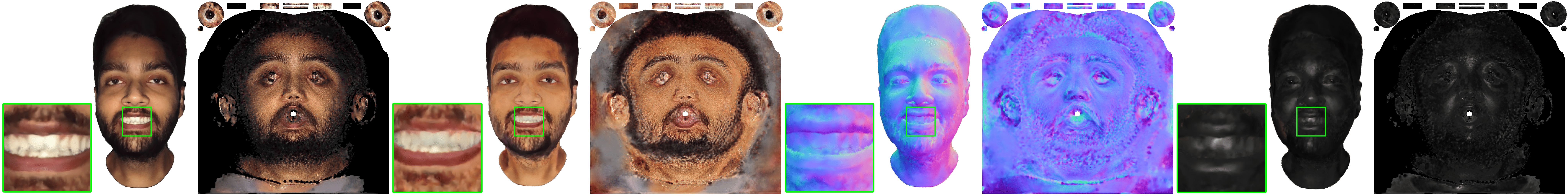}
    \caption{Decomposition of appearance \& geometry in UV space.}
    \label{fig:qual_ours}
    \vspace{-3mm}
\end{figure*}
% \begin{table}
% \caption{Advantages of our textured head representation}
% \centering
% \begin{tabular}{|l|c|c|c|}
% \hline
% \textbf{Features} & \textbf{Pt.Avatar} & \textbf{GBS} & \textbf{Ours} \\
% \hline
% Editing & \ding{55} & \ding{55} & \ding{51} \\
% \hline
% PBR & \ding{55} & \ding{55} & \ding{51} \\
% \hline
% Realtime Rendering & \ding{55} & \ding{51} & \ding{51} \\
% & ($5$ FPS) & ($100$+ FPS) & ($40$ FPS) \\
% \hline
% Low VRAM & \ding{55} & \ding{51} & \ding{51} \\
% & ($\sim$40 GB) & ($\sim$14 GB) & ($\leq$8GB) \\
% \hline
% % Low Storage & \ding{55} & \ding{55} & \ding{51} \\
% % & ($300$ MB) & (2GB) & ($\leq$50MB) \\
% % \hline
% % \multicolumn{4}{l}{$\dagger$ Not physically based}\\
% \end{tabular}
% \label{tab:features}
% \end{table}
% 
% 
% 
% \begin{table}
% \caption{Advantages of the proposed textured head representation}
% \centering
% \begin{tabular}{|l|c|c|c|c|}
% \hline
% \textbf{} & \textbf{Edit} & \textbf{PBR} & \textbf{Realtime} & \textbf{Low VRAM} \\
% \hline
% \textbf{Pt.Avtr.} & \ding{55} & \ding{55} & \ding{55} ($5$ FPS) & \ding{55} ($\sim$40 GB) \\
% \hline
% \textbf{GBS} & \ding{55} & \ding{55} & \ding{51} ($100$+ FPS) & \ding{51} ($\sim$14 GB) \\
% \hline
% \textbf{Ours} & \ding{51} & \ding{51} & \ding{51} ($140$ FPS) & \ding{51} ($\leq$8GB) \\
% \hline
% \end{tabular}
% \label{tab:features}
% \end{table}
% 
% 
\section{Related Works}
\noindent
\textbf{Lightstage Capture Systems:}
Polarization has long been used to decompose scene illumination\cite{Mller1995PolarizationBasedSO, 990468, 85655}, leveraging the fact that the single bounce specular reflection does not alter the polarization of the incoming light. Using polarization filters and controlled lighting, Debevec et al.\cite{10.1145/344779.344855} pioneered the first Lightstage system to estimate human face reflectance, efficiently capturing how the face appears when lit from every possible lighting direction. Though the initial setup was meant for static capture, subsequent advancements\cite{10.1145/2070781.2024163, 10.1145/1731047.1731055, Woodham1980PhotometricMF} proposed techniques to compensate for motion, expanding the capture area \& improving the surface fidelity. \cite{10.1145/3272127.3275073} proposed a multi-view setup for dynamic facial texture acquisition without the need for polarized illumination. Following this, \cite{10.1145/3386569.3392464} reintroduced polarization without active illumination to model subsurface-scattering. Recently proposed \cite{10.2312:egs.20221019} proposed further improvements in the system to include global illumination and polarization modeling. Though these volumetric lightstage-based solutions for capturing face reflectance \& geometry deliver impressive visual results, they rely on sophisticated hardware, making them highly expensive and bulky, while requiring a significant level of expertise to operate. Hence, they face challenges in scaling to large numbers of identities.\\
\\
\textbf{Data-driven Personalized Head Avatars:}
Facial expression control has advanced from blendshape-based methods in VFX and gaming to data-driven techniques that estimate statistical bases from 3D head scans, enabling nuanced head pose and expression control~\cite{10.1145/311535.311556, Pighin1998SynthesizingRF, Cao2015RealtimeHF, multilinearvlasic, DBLP:journals/corr/abs-1807-10267, DBLP:journals/corr/abs-1804-03786}. However, models like FLAME~\cite{FLAME:SiggraphAsia2017} often miss subtle expressions and detailed, person-specific geometry (e.g., hair, beard, skin deformations). For photorealistic avatars, several methods use differential rendering from multiview videos in an analysis-by-synthesis framework~\cite{Gafni_2021_CVPR, kirschstein2023nersemble, zhao2023havatar, kirschstein2023diffusionavatars, qian2023gaussianavatars, giebenhain2024npga, xu2023gaussianheadavatar}. While NeRF-based approaches yield personalized avatars, they suffer from slow rendering and coarse geometry, motivating recent methods to employ 3D Gaussian splats (3DGS) \cite{kerbl3Dgaussians} for efficient real-time performance—albeit with illumination baked into the appearance, which prevents relighting. A few works~\cite{saito2024rgca, li2024uravatar} achieve relightable avatars using lightstage data, but their dependence on such systems limits demographic diversity. Diffusion-based techniques~\cite{10.1145/3610543.3626169, srivastava2024wordrobetextguidedgenerationtextured, chen2023text2tex} and neural editing methods~\cite{haque2023instructnerf2nerfediting3dscenes, mendiratta2023avatarstudiotextdrivenediting3d} enable texture edits yet remain slow and struggle with novel poses. A recent approach~\cite{wang2024mega} uses neural texture maps for flexible editing but also relies on optimization-heavy processes. In this paper, we propose an affordable, scalable method for creating animatable, relightable, and editable head avatars.\\

\noindent
\textbf{Head Avatars from Monocular Video(s):} Creating animatable head avatars from monocular videos is yet another interesting \& challenging direction. Several existing learning-based methods aim to eliminate the need for multiview information by tracking dynamically evolving head topology over a monocular video sequence, either in the form of a parametric head mesh \cite{baert2024spark} or a neural head representation \cite{giebenhain2024mononphm}. For more detailed and expressive monocular head avatars, existing methods propose to use primitives, such as points\cite{Zheng2023pointavatar} or 3D Gaussians\cite{chen2023monogaussianavatar}, on top of FLAME\cite{FLAME:SiggraphAsia2017} model.  Methods like \cite{shao2024splattingavatar} \& Gaussian Blendshapes\cite{ma2024gaussianblendshapes} attempt to use Gaussian splats as their primitives, but owing to the nature of ill-defined normals of 3D Gaussians, lack high-quality surface details. These methods also fail to provide any editability support to head meshes. PointAvatar\cite{Zheng2023pointavatar}
aims to reconstruct a relightable human head avatar in a highly constrained fixed illumination setting. While it achieves impressive rendering quality, the relighting is not Physically-based, and does not take into account the properties of skin reflectance. Moreover, points as primitives are memory intensive in nature, making \cite{Zheng2023pointavatar} unsuitable for realtime rendering. Furthemore, none of the existing monocular head avatar methods directly provides texture maps to enable flexible \& quick editing appearance editing. Though, FlashAvatar\cite{xiang2024flashavatarhighfidelityheadavatar} embeds 3D Gaussians within FLAME's UV space for initialization, it doesn't learn any UV texture maps and lacks relighting support.
% The use a deformed flame mesh for facial features and a Gaussian representation to allow appearance editing, but owing to the use of multiple cameras, they are not scalable, and cannot provide realtime 
% rendering.
% {\color{red}check if this is true?}. 
% Instead, we utilize Gaussians to represent the whole head and allow editing directly into the Gaussian head avatar representation.
% Though some methods\cite{} attempt to disentangle shading information from the geometry by modelling pose-dependent lighting effects.

% 
% 
\section{Method}
\label{sec:Methodology}
To create affordable person-specific relightable head avatars in a monocular setting, we propose an effortless polarized data acquisition process. 
% to capture cross-polarized \& parallel-polarized monocular video sequences using a tripod-mounted smartphone, equipped with polaroid filters. 
The captured polarized data is then used to create a novel textured head avatar representation. In order to control head pose \& expressions, we propose to use FLAME, combined with 2DGS\cite{Huang2DGS2024} to model person-specific offset details \& appearance. For relighting, we propose a novel self-supervised learning scheme to decompose the polarized information into appearance and shading information in the form of albedo, normal \& roughness UV maps in FLAME's parametrization space (as shown in \autoref{fig:qual_ours}), while also learning a environment cubemap. Additionally, to model dynamic changes in the appearance and geometry due to pose/expressions, we learn expression-dependent residual UV maps.
% via an expression-conditioned hash-encoded MLP. The parameters of the MLP and underlying $uv$-maps are optimized by associating each 2D Gaussian with a $uv$-coordinate, and performing differentiable rendering by splatting the Gaussians onto the polarized video frames. Once training is converged, we can perform appearance editing by directly altering the base albedo map's textures as we desire.
We now describe each component detail. 
%We now describe our polarized data capture process, proposed head avatar representation and self-supervised training methodology in detail.
% for learning attribute $uv$-maps in detail. 

% We propose a dynamic polarized data capture process by equipping back cameras and flashlight of a commodity smartphone with polarization filter

% To generate a relightable head avatar, we first capture a dual-view RGB video, with polaroid filters, to disentangle the specular details from albedo information, followed by gaussian splat-based reconstruction.

%-------------------------------------------------------------------------

\subsection{Polarized Dynamic Data Capture}
\label{sec:capture}
%\vjnote{Capture is probably better word than acquisition for the section heading.}
In order to capture surface illumination information, \cite{azinovic2022polface} proposed a static smartphone-based capture setup, utilizing a single back camera with polaroid filter to sequentially capture several cross and parallel-polarized images of a person one-at-a-time, demanding the person to remain stationary for an extended period. However, expanding this setup to a dynamic scenario is non-trivial and requires careful consideration to achieve a scalable and calibration-free capture process. In order to achieve this, we mount a single smartphone on a tripod in a dark room with its flashlight as the only source of illumination (point light) to avoid estimation of complex scene lighting. As shown in \autoref{fig:setup}, we cover the flashlight using a thin linearly polarized filter/film (highlighted in purple), about $4.8$ microns in thickness. We also cover the two back cameras of the smartphone with the same polarized film, with a relative angle difference of $0^\circ$ (red) and $90^\circ$ (green) w.r.t. the flashlight's film, essentially making one camera parallel-aligned and the other one cross-aligned to the flashlight's polarization. The usage of polaroid filters introduces a tint-shift between the capture stream of the two cameras as can be seen in \autoref{fig:setup}. While \cite{azinovic2022polface} precomputes an approximate affine color correction matrix, we delegate the tint-shift correction during the training as discussed in \autoref{sec:optimization}, to make the capture process seamless. % \vjnote{We are probably mentioning seamless or effortless too many times?}.
% talk about point-light assumption and attenutation map in discussion, [TODO]
% write about dual camera baseline, [TODO]

% Using the proposed setup, 
% Using the aforementioned setup, 
Using this setup, we simultaneously record two uninterrupted monocular videos (cross \& parallel-polarized) of a human subject enacting a predefined set of diverse expressions \& poses to ensure adequate deformations \& lighting information from different angles is captured. Both the video streams are captured in $1920\times1080$ resolution at $24$ FPS. It is important to note that, though we use two back cameras at the same time to ensure both the video streams are in sync, we do not use any stereo-based depth or alignment information due to negligible baseline distance between the two cameras, thereby facilitating avatar creation by treating the two streams as separate monocular RGB videos. 
% \textit{Please refer to the supplementary video to see our capture process in action.}
% 
% 
% Our capture process uses a single phone with dual back camera (wide angle and normal) as its sole capture device. Both of the lenses are covered with linear polaroid filters, with the relative angle of polarisation between them equal to $90^\circ$. The flashlight is also covered with a linear polaroid filter, aligned in its angle of polarisation to one of the lenses, as in (??). The stream from both the rear lenses is captured simultaneously, while zooming and cropping on the stream from the wide-angle lens, to alleviate distortions and bring the effective focal length closer.
% 
% We utilize a stereo video stream flame optimization process, in which we jointly optimize focal length and camera poses of the two streams, for a common set of flame parameters per frame. We assume the two camera poses to be the same and optimize only focal lengths for the two camera streams.
% 
% 
%-------------------------------------------------------------------------

\subsection{Textured Gaussian Head Representation}
\label{sec:head_rep}
Given the cross-polarized and parallel polarized monocular video sequences, we employ the head-tracking~\cite{qian2024versatile} proposed in \cite{qian2024gaussianavatars} to obtain the subject-specific FLAME head mesh $\mathcal{M}$ temporally tracked separately over both monocular sequences. We propose to model the remaining subject-specific details on top of FLAME mesh using 2D Gaussian Splats (2DGS)\cite{Huang2DGS2024}, which are essentially `flat' 2D planar elliptical disks embedded in 3D space. 2DGS uses an explicit ray-splat intersection technique, resulting in a perspective-correct splitting and more accurate surface reconstruction. Additionally, this enables direct surface regularization through normal constraints\cite{Huang2DGS2024}, improving the quality of normals for shading.
\begin{figure}[t]
    \centering
    \includegraphics[width=\linewidth]{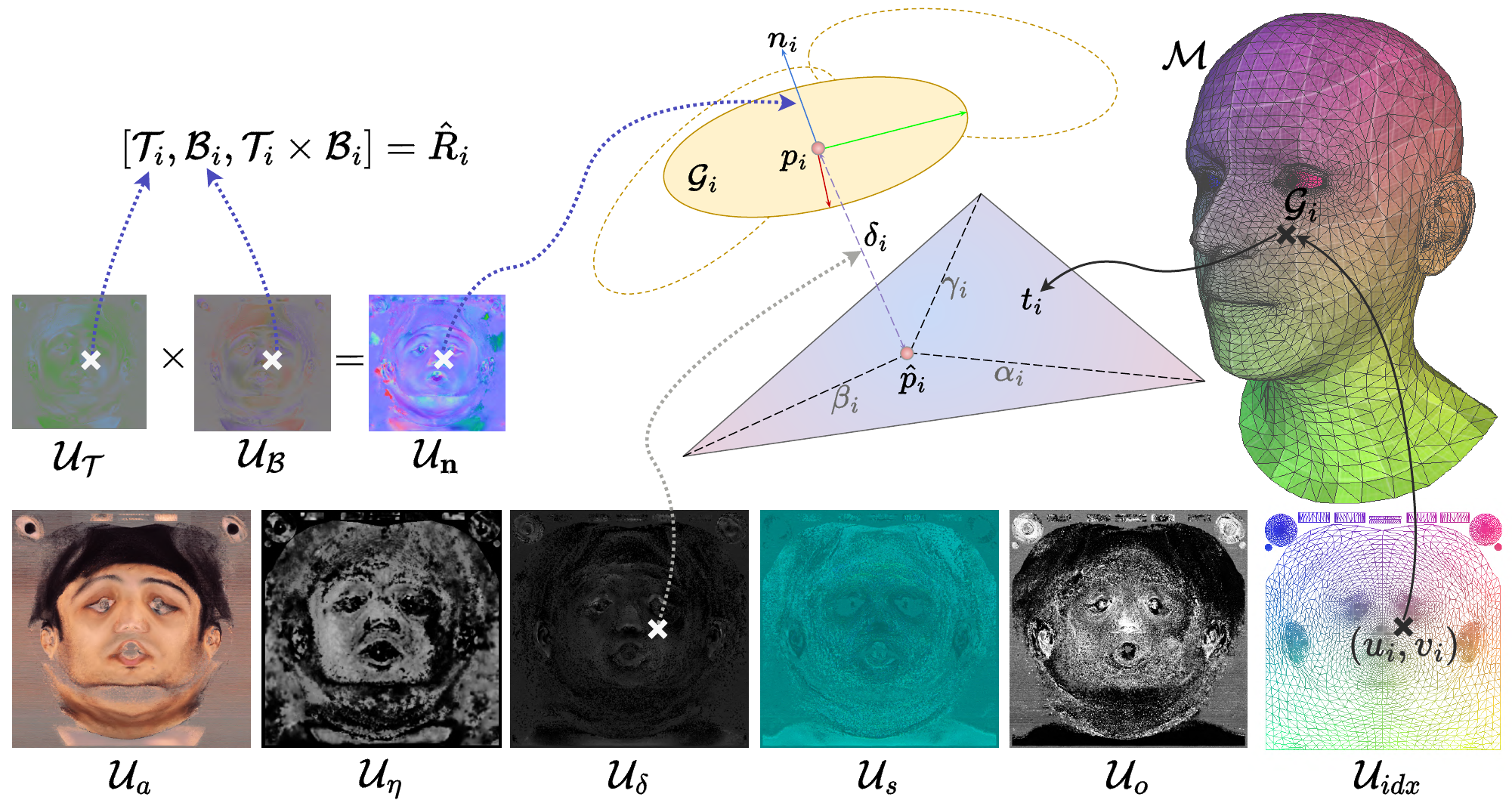}
    \caption{Proposed textured Gaussian head representation with primary UV attribute maps.}
    \label{fig:representation}
\end{figure}
% 
% To allow for easier editing and manipulation of head avatar appearance, we learn the appearance of Gaussian splats over several texture maps, namely, an albedo map, a normal map, a roughness map and an offset map. 
\noindent
Each 2D Gaussian disk $\mathcal{G}_{i}$ is characterized using its mean position $p_{i}\in \mathbb{R}^3$ in 3D space, its 2D scale $S_i \in \mathbb{R}^2$, a quaternion $q_i \in \mathbb{R}^4$ defining its orientation, and its opacity $o_i \in \mathbb{R}^1$. Similar to 3DGS~\cite{kerbl3Dgaussians} , the factorized covariance matrix $\Sigma_i$ for each 2D Gaussian $\mathcal{G}_{i}$ is defined as 
% \begin{equation}
   $\Sigma_i=R_i S'_i {S'_i}^\top R_i^\top$
% \end{equation}
, where $R_i$ is the rotation matrix, derived from the 6D rotation vector $r_i$, and $S'_i=[S_i,0]^\top$ $\in \mathbb{R}^3$ (i.e. the third dimension of 3D scale vector is set to zero to represent a flat 2D Gaussian in 3D space).
%\vjnote{This seems to overly introducing the notations. Rotations are represented in 3 forms: q, r and R. Only introduce the necessary ones. Remove q and r unless used elsewhere.}
% 
% Our representation also utilizes 2DGS because of their well-defined normals and better normal regularization, but instead of defining the Gaussian in 3D object space [check terminology??], 
% As shown in \autoref{fig:representation}, , we define it in the tangent space of the parametric head mesh $\mathcal{M}$. This strategy helps us 
However, we propose to associate all the learnable attributes of 2D Gaussians to FLAME's UV space, by embedding the Gaussians in tangent space of the triangulated mesh, instead of 3D object space.
% In other words, instead of initializing the mean position $p_{i}$ of each 2D Gaussian in 3D space as $(x_i,y_i,z_i)$, 

\vspace{1mm}
\noindent
\textbf{Embedding 2D Gaussians in UV space:}
We begin with initializing a 2D Gaussian $\mathcal{G}_i$ for a UV-coordinate $(u_i,v_i)$ in FLAME's canonical UV space $\mathcal{U}$, as shown in \autoref{fig:representation}. We also estimate a face index map, $\mathcal{U}_{idx}$, to store the triangle (face) index $m$ on which a UV-coordinate of $\mathcal{U}$ lies (undefined for empty UV space); this is a one-time computation. To transform Gaussians from canonical pose/expression to the deformed pose/expression directly via posed $\mathcal{M}$, we first compute the associate face index $m = \mathcal{U}_{idx}(u_i,v_i)$, which gives us a triangle $t_m = (\mathrm{v}_{m_0},\mathrm{v}_{m_1},\mathrm{v}_{m_2}) \in 3 \times \mathbb{R}^3$ (3D positions of $t_m$'s vertices). For each Gaussian $G_i$, we compute the initial mean position in the posed 3D space as, $\hat{p}_i =(\alpha_i*\mathrm{v}_{m_0}) + (\beta_i*\mathrm{v}_{m_1}) + (\gamma_i*\mathrm{v}_{m_2}) $, where $\alpha_i, \beta_i, \gamma_i$ are the barycentric coordinates for the point $(u_i, v_i)$ \& $\alpha_i+\beta_i+\gamma_i=1$. This formulation enables us to define the remaining attributes of the Gaussians (orientation, scale, offsets, appearance, etc.) as learnable UV-maps.\\
% To embed pose/expression information in the UV space, we use the LBS-based displacement of FLAME mesh vertices from the canonical pose and store them in UV space as $\mathcal{U}_{LBS}(u_i,v_i)=\hat{d}_i=(\alpha_i*\hat{d}_{m_0}) + (\beta_i*\hat{d}_{m_1}) + (\gamma_i*\hat{d}_{m_2})$, where $\hat{d}_{m_j}$ is the relative displacement of $\mathrm{v}_{m_j}$ in posed 3D space from it's 3D position in canonical (unposed) space. [TODO: Move to training]

\noindent
\textbf{Gaussian Attributes as UV Maps:}
Given the aforementioned formulation, we now explain how we model the Gaussian attributes as texture maps. We sample 2D Gaussians for every valid UV-texel (excluding empty space) and define a set of \textit{primary} UV-maps for each attribute$-$$\mathcal{U}_a$(albedo map), $\mathcal{U}_\eta$ (roughness map), $\mathcal{U}_\mathcal{T}$ (tangent map), $\mathcal{U}_\mathcal{B}$ (bitangent map), $\mathcal{U}_\delta$ (offset map), $\mathcal{U}_s$ (scale map) \& $\mathcal{U}_o$ (opacity map). As shown in \autoref{fig:representation}, for each Gaussian $G_i$ with its own UV-coordinate, we can query any attribute $\Omega_i=\mathcal{U}_{\Omega}(u_i,v_i)$, where $\Omega_i$ can be albedo color $a_i\in {\mathbb{R}^2}$, roughness $\eta_i\in {\mathbb{R}}$, tangent vector $\mathcal{T}_i\in {\mathbb{R}^3}$, bitangent vector $\mathcal{B}_i\in {\mathbb{R}^3}$, 2D scale $s_i\in {\mathbb{R}^2}$, and opacity $o_i\in {\mathbb{R}}$. To estimate orientation for the 2D Gaussian $G_i$, we first query tangent $\mathcal{T}_i=\mathcal{U}_\mathcal{T}(u_i,v_i)$ \& bitangent $\mathcal{B}_i=\mathcal{U}_\mathcal{B}(u_i,v_i)$ and perform Gram–Schmidt orthogonalization. The tangent space rotation for Gaussian $G_i$ is defined as $\hat{R}_i = [T_i, B_i, T_i \times B_i] \in \mathbb{R}^{3\times3}$. We then obtain the 3D world space rotation/orientation 
% 
% We then use precomputed TBN matrix\cite{?} $\mathbb{T}$ to transform the $\hat{R}_i$ to 3D object space as: 
% \begin{equation}
$ R_i=\mathbb{T}*\hat{R}_i$
% \end{equation}
% 
, where $\mathbb{T}$ is a $3\times3$ transformation matrix for transforming a vector from tangent space to world space. Additionally, we compute the normal vector $\mathcal{N}_i=\mathcal{T}_i \times \mathcal{B}_i$, which is required for shading.
To account for geometrical details far from the surface of the parametric head mesh $\mathcal{M}$, we define offset $\delta_i=\mathcal{U}_\delta(u_i,v_i)$ along the normal direction and shift the initial mean position $\hat{p}_i$ of the 2D Gaussian in 3D space as follows:
\begin{equation}
\label{eq:shifted_position}
    {p}_i = \hat{p}_i + \xi_i*\mathbb{T}*\delta_i
\end{equation}
\noindent
where ${\xi}_i$ is the area-adjustment factor, which restricts the Gaussians to go too far from their associated triangles or grow too big in scale, and is computed as:
\begin{equation}
    \xi_i = (\alpha_i*a_{m_0}) + (\beta_i*a_{m_1}) + (\gamma_i*a_{m_2})
\end{equation}
\begin{equation}
    a_{m_j}=\dfrac{1}{|\mathcal{N}(\mathrm{v}_{m_j})|}\sum_{k\in\mathcal{N}(\mathrm{v}_{m_j})}\sqrt{Area(t_k)}
\end{equation}
here, $\alpha$, $\beta$, $\gamma$ are barycentric coordinates, and $a_{ij}$ for vertex $v_{ij}$ is the mean-root-sum of area of triangles in $v_{m_j}$'s neighborhood $\mathcal{N}(v_{m_j})$. 
% $\xi$ restricts the Gaussians associated with smaller triangles to go too far or grow too big in scale. \\
% 
% computed as:
% \begin{equation}
%     \xi_i = (\alpha_i*a_{m_0}) + (\beta_i*a_{m_1}) + (\gamma_i*a_{m_2})
% \end{equation}
% \begin{equation}
%     a_{m_j}=\dfrac{1}{|\mathcal{N}(\mathrm{v}_{m_j})|}\sum_{k\in\mathcal{N}(\mathrm{v}_{m_j})}\sqrt{Area(t_k)}
% \end{equation}
% here, $\alpha$, $\beta$, $\gamma$ are barycentric coordinates, and $a_{ij}$ for vertex $v_{ij}$ is the mean-root-sum of area of triangles in $v_{m_j}$'s neighborhood $\mathcal{N}(v_{m_j})$. $\xi$ restricts the Gaussians associated with smaller triangles to go too far or grow too big in scale. \\

\noindent
\textbf{Residual UV Maps:} In addition to primary UV-maps, we also maintain a set of UV-maps to store expression-dependent residuals. We define $k$ residual UV-maps, where each $(u_i, v_i)$ coordinate stores residuals $\Delta \Omega_i $ on top of primary attributes $\Omega_i $ to model different geometrical and appearance changes observed throughout the sequence. Specifically, we store $\Delta p_i, \Delta a_i, \Delta \mathcal{T}_i, \Delta {B}_i, \Delta s_i, \Delta \eta_i$. In order to provide expression guidance, we take expression parameters of FLAME head mesh, $\psi\in \mathbb{R}^{100}$, and project them onto a k-dimensional space using a projection matrix, $\bf{\Pi} \in \mathbb{R}^{k\times 100}$, to obtain linear blend weights $\mathcal{W} = \bf{\Pi}\cdot\psi \in \mathbb{R}^k$. The final blended residuals for a Gaussian $G_i$ are obtained as follows:
\begin{equation}
\label{eq:blending}
    \Delta\Omega_i = \mathcal{U}_{\Delta}(u_i, v_i) = \sum_{w_j\in\mathcal{W}} w_j*\mathcal{U}_{\Delta_j}(u_i,v_i)
\end{equation}
\noindent
The residuals are added to primary attributes $\Omega_i$ to obtain final attributes $\Omega'_i$ using the following equation:
% \begin{equation}
%     % \hat{\Omega}_i = \Omega_i * exp(\Delta\Omega_i)
%     \
% \end{equation}
% 
\begin{equation}
\label{eq:add_residuals}
\Omega'_i =
\begin{cases} 
\Omega_i + \Delta\Omega_i & \text{; for } p_i, \mathcal{T}_i, \mathcal{B}_i \\ 
\Omega_i * exp(\Delta\Omega_i) & \text{; for } a_i, s_i, \eta_i \\
\Omega_i & \text{; for } o_i .
\end{cases}
\end{equation}

\begin{figure*}[h]
    \centering
    \includegraphics[width=\linewidth]{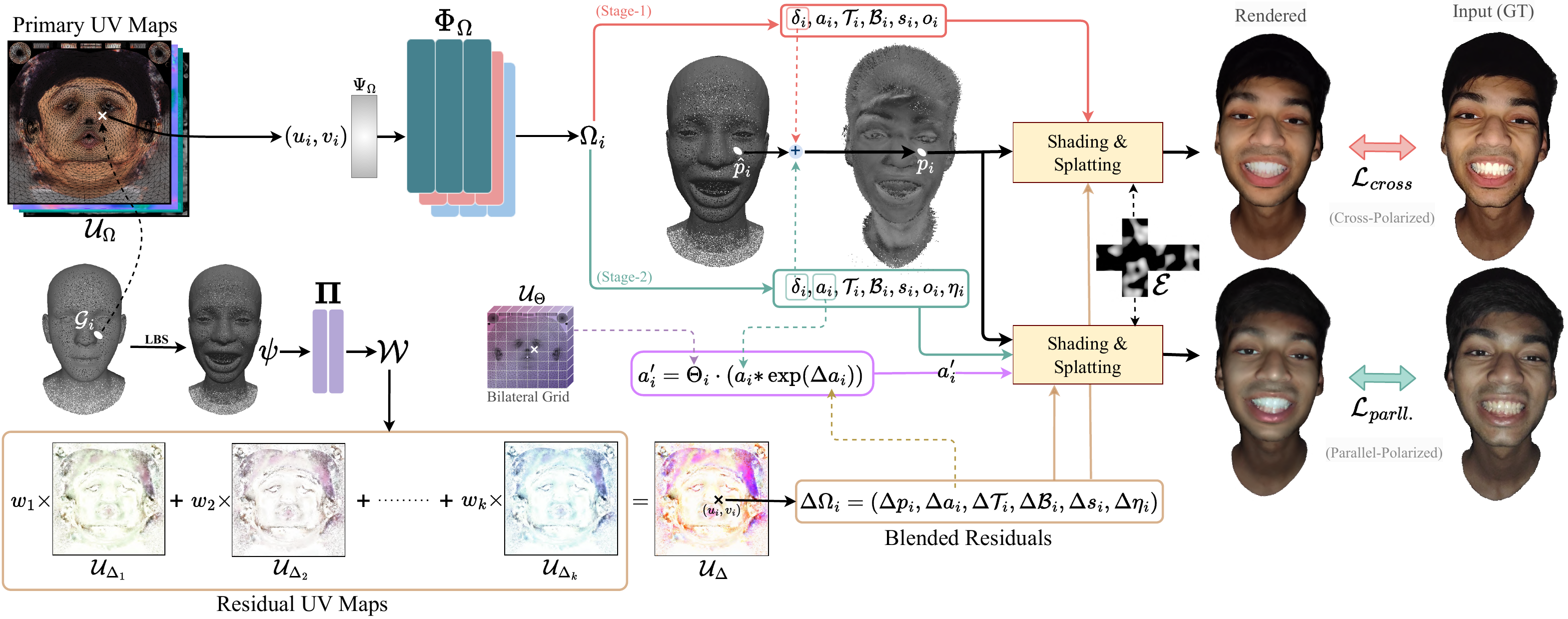}
    \caption{Proposed two-stage training strategy to learn textured Gaussian head avatars with decomposed appearance and geometry.}
    \label{fig:training}
\end{figure*}
\subsection{Learning Facial Geometry \& Reflectance}
\label{sec:optimization}
\noindent
\textbf{Rendering Equation \& BRDF:} 
For appearance modelling, the original 2DGS uses Spherical Harmonics\cite{Fridovich-Keil_2022_CVPR} to learn the appearance of each 2D Gaussian. Though SH-coefficients allow for an accurate view-dependent appearance for static scenes, 
% they cannot be reliably used for a dynamically evolving head geometry, as it is non-trivial to transform the underlying SH-bases according to the parametric head deformation. Additionally,
higher-order SH-coefficients are needed to model high-frequency details (e.g. surface normals and roughness), exponentially increasing the memory requirements. More importantly, appearance editing is not directly possible when using such representation. This motivates us to replace  SH-coefficients with a single RGB color for appearance $c_i$. For view-dependent color, we implement a new Physically-based SVBRDF\cite{1965ApOpt...4..767N} shader for 2DGS, to estimate the Gaussian's appearance given the viewing direction `$\mathbf{\omega}$' and light direction `$l$' from a point source (flashlight). Furthermore, for learning shading to enable relighting, we disentangle the appearance $c_i$ into albedo, specular and diffused components, namely $a_i$, ${f_s}_i$ \& ${f_d}_i$. For specular component ${f_s}_i$, we use Cook-Torrance\cite{10.1145/357290.357293} microfacet specular BRDF defined as follows:
% 
% replace the sh coefficients with single albedo value $a_i$. Therefore,
% 
\begin{equation}
{f_s}_i(l,\mathbf{\omega},\eta_i) = k_s*\dfrac{D(h)F(\mathbf{\omega}, h)G(l, \mathbf{\omega}, h)}{4(\mathrm{n} \cdot l)(\mathrm{n} \cdot \mathbf{\omega})}\\
\end{equation}
where, $h$ is the half vector, bisecting the angle between $l$ \& $\mathbf{\omega}$; $k_s$ is a constant specular gain. 
Unlike \cite{azinovic2022polface}, which uses a spatially varying specular gain $k_s$ (which is a learnable constant in our case), we instead use spatially varying roughness $\eta_i$ associated with each Gaussian $\mathcal{G}_i$ to handle difference in speculariy of skin, hair, teeth etc.
% 
% unlike spatially varying specular gain used by \cite{azinovic2022polface}, i.e. $k_s$ is same for all Gaussians.
% (different values for skin, hair \& teeth region).
% 
% \textcolor{cyan}{(SHOULD WE EXPLAIN WHY?)}. 
Following \cite{azinovic2022polface}, we use Schlick's approximation\cite{https://doi.org/10.1111/1467-8659.1330233} for the Fresnel term $F$. For the NDF term $\mathbf{D}$, we use an alternative approximation proposed by Trowbridge-Reitz\cite{10.5555/2383847.2383874}:

% \textcolor{cyan}{(TODO: CONFIRM \& CITE THESE APPROXIMATIONS)}We expand the NDF term\cite{} and geometric term as follows:
\begin{equation}
D(h,\eta_i) = \dfrac{{\eta_i}^2}{\pi((\mathrm{n} \cdot h)^2({\eta_i}^2 - 1) + 1)^2}
\end{equation}
% 
% \begin{equation}
% G(l,\mathbf{\omega},h) = G_{Shlick}(l)*G_{Shlick}(\mathbf{\omega}) = [G_{Shlick}(\mathbf{\omega})]^2
% \end{equation}
% 
% \begin{equation}
% G(l,\mathbf{\omega},h) = {G^2}_{Schlick}(\mathbf{\omega}) = \left[ \dfrac{\mathrm{n \cdot \omega}}{(\mathrm{n \cdot \omega})(1-\lambda)+\lambda}\right]^2
% \end{equation}

\noindent
Our geometric term $\mathbf{G}$ uses Smith's variant of Shlick-GGX approximation\cite{https://doi.org/10.1111/1467-8659.1330233}.  Please note that similar to \cite{azinovic2022polface}, we assume that $l=\mathbf{\omega}$ since the flashlight is very close to the camera lens, making the implementation straightforward. 

\begin{equation}
G(l,\mathbf{\omega},h) = G_{Shlick}(l)*G_{Shlick}(\mathbf{\omega}) = [G_{Shlick}(\mathbf{\omega})]^2
\end{equation}
\begin{equation}
G(l,\mathbf{\omega},h) = {G^2}_{Schlick}(\mathbf{\omega}) = \left[ \dfrac{\mathrm{n \cdot \omega}}{(\mathrm{n \cdot \omega})(1-\lambda)+\lambda}\right]^2
\end{equation}
where $\lambda=\eta_i/2$ for remapping Schlick-GGX to match with Smith's formulation\cite{smith-ggx}.
% 
% 
% % 
% 
% 
\\
\\
\noindent
For the diffuse component, we use the BRDF model proposed by Ashikhmin \& Shirley \cite{shirley}:
\\
\begin{equation}
\label{eq:diffused_color}
{f_d}_i({a}_i, \mathbf{\omega}) = \dfrac{28{a}_i}{23\pi}(1 - F_0)(1 - (1 - \frac{\mathrm{n}^\top\mathbf{\omega}}{2})^5)^2,
\end{equation}
where $a_i$ is the albedo color and $F_0$ = 0.04 is the reflectance of the skin at normal incidence. Besides shading using point light, we also incorporate ambient shading component $f_{env}$ to account for the light bouncing around the environment. We follow the differentiable version of the \textit{split sum} shading model proposed in \cite{Munkberg_2022_CVPR} to learn environment lighting from image observations through optimization. The final shaded color is computed as:
\begin{equation}
\label{eq:shaded_color}
c_i=f{_d}_i+{f_s}_i+f_{env} .
\end{equation}
\autoref{fig:training} illustrates our novel training methodology for learning a textured Gaussian head avatar. Given cross-polarized and parallel-polarized video sequences, we learn the aforementioned Gaussian attributes in two separate stages.  We employ a set of \textit{stacked} MLPs, $\Phi_\Omega$, which consists of several hash-encoded MLPs\cite{mueller2022instant} with hash-encoding $\Psi_\Omega$, to predict UV attributes $\Omega_i = \mathcal{U}_\Omega(u_i, v_i)$ for a 2D Gaussian $G_i$. 
% For pose-specific albedo, we employ another hash-encoded MLP $\Phi_\Delta$ with hash encoding $\psi_\Delta$. 
% For expression-based residual maps, we initialize random
For residual UV-maps, we initialize learnable UV-tensors $\mathcal{U}_{\Delta_j}$ with $z=|\Delta \Omega_i|$ channels, where $j=1$ to $k$.
To handle the global tint-shift between the cross-polarized and parallel-polarized frames, we use a low-dimensional Bilateral Grid\cite{wang2024bilateralguidedradiancefield} $\mathcal{U}_\Theta$ with learnable affine transformation matrix $\Theta_i$ for each UV-texel.  Additionally, since we allow $(u_i,v_i)$ to optimize over time, we also precompute the triangle-index map $\mathcal{U}_{idx}$ for querying the triangle index from UV-coordinates. Finally, we define a learnable cube-map $\mathcal{E}$ to store the environment lighting information. All the values specified within $\mathcal{U}_\Omega$, $\mathcal{U}_{\Delta_j}$, $\mathcal{U}_\Theta$, $\bf{\Pi}$ and $\mathcal{E}$ are initialized at random and optimized via differential-rendering losses in a self-supervised manner, using cross-polarized sequence in the first stage and parallel-polarized sequence in the second.

 % We begin with initializing per-$uv$-texel 2D Gaussians $G_i$ with $uv$-coordinate $(u_i,v_i)$ to learn their attributes embedded within the UV space. 
 During the first stage, we focus on learning $\mathcal{U}_\Omega$ (except the roughness map $\mathcal{U}_\eta$), along with $\mathcal{U}_{\Delta_j}$. We feed $(u_i,v_i)$ as input to hash-encoded MLP $\Phi_\Omega$ to predict $\Omega_i = \Phi_{\Omega}(\Psi_\Omega(u_i,v_i))$, i.e. albedo $a_i$, tagent $\mathcal{T}_i$, bitangent $\mathcal{B}_i$ , scale $s_i$, opacity $o_i$, and estimate shifted mean position $p_i$ (\autoref{eq:shifted_position}). For a given pose/expression $\theta$, we compute linear blending weights $\psi = \bf{\Pi}\cdot\theta$ and use \autoref{eq:blending} to obtain $\Delta\Omega_i$. We then add the residuals to the primary attributes using \autoref{eq:add_residuals}. At last, we compute final shaded color $c_i$ via \autoref{eq:shaded_color}(ignoring the specular component) and use it along with $\mathcal{T}_i$, $\mathcal{B}_i$, $s_i$, $o_i$ to perform 2D Gaussian splatting to rendered the image \& minimize loss $\mathcal{L}_{cross}$.

During the second stage, we freeze the optimization of primary \& residual UV-maps ($\mathcal{U}_{a}$ \& $\mathcal{U}_{\Delta a}$), and focus on learning roughness only. We also learn bilateral grid $\mathcal{U}_\Theta$ to account for the tint shift between the albedo learned from cross-polarized images (in the previous stage) and parallel-polarized images. We obtain tint-corrected albedo as: 
    \begin{equation}
    \label{eq:tint}
     a'_i = \Theta_i \cdot (a_i*\exp(\Delta a_i)).
    \end{equation}
We include both diffuse and specular components during second stage while computing shaded color $c_i$ (\autoref{eq:shaded_color}). We query the remaining attributes learned in the first stage and obtain a rendered image via splitting and minimize the loss  $\mathcal{L}_{parll.}$
We define both $\mathcal{L}_{cross}$ and $\mathcal{L}_{parll.}$ as:
\begin{equation}
\label{eq:loss_function}
    \mathcal{L}_{cross/parall.} = \mathcal{L}_1 + \mathcal{L}_{SSIM} + \mathcal{L}_{LPIPS} + \mathcal{L}_{scale}
\end{equation}
where, $\mathcal{L}_{scale}=\sum_{i}\mathrm{max}(0, s_i-0.3)$ to prevent Gaussians to grow too large.

In both the stages, we initialize the environment cube map $\mathcal{E}$ at random and optimize its values via \autoref{eq:shaded_color}.
During inference, for efficient rendering, we discard the MLP $\Phi_\Omega$ and use attributes stored in the primary UV-maps $\mathcal{U}_\Omega$, while relying on $\bf{\Pi}$ \& residual UV-maps $\mathcal{U}_\Delta$ to handle poses/expressions-dependent deformations.
% We also provide a choice to even discard $\Phi_\Delta$ by storing $\mathcal{U}_{\Delta a}$ for every frame, and using LBS to extrapolate within training poses for even faster relighting/editing of seen poses/expression.

\section{DuoPolo Dataset}
We also introduce \textbf{``DuoPolo''} dataset, which consists of cross-polarized \& parallel-polarized video sequences of $10$ subjects, captured using the proposed capture setup. Each data sample consists of a $90s$ video sequence of a human subject enacting diverse head poses and facial expressions, followed by a short utterance of a phrase (to capture subtle lip motions during talking). Each sequence is captured in $1920\times1080$ resolution at 24 FPS. Along with the video, we also provide per-frame background/foregound segmentation mask, 3D facial landmarks, and parametric head mesh, FLAME\cite{FLAME:SiggraphAsia2017}, tracked over both the videos using \cite{qian2024versatile}. The proposed dataset is first-of-its-kind which aims to make polaroid facial performances accessible to everyone, bridging the gap between expensive light-stage data and affordable head avatars. 
% As a future use case, we plan to  one can easily scale the dataset to include a large number of diverse demographics which can be further used to train a lower quality, yet highly generalized universal prior model as compared to \cite{}. [TODO: Move to conclusion]

\section{Training \& Implementation Details}
 We learn $1024\times 1024$ size texture maps via Hash encoded Coordinate MLPs . The Hash-encoded MLP for texture map synthesis has 1 hidden layer with 16 neurons. The Hash Table has $7$ levels, each with a size $2^{20}$. The base resolution of the hash table was $512$, with a growth factor of $1.26$. The pose-dependent MLP had 2 hidden layers, with $32$ neurons each. The Hash grid for the pose-dependent MLP had 16 levels, each of size $2^{18}$ with a base resolution of $16$ and a growth factor of $1.4$. Both the MLP are trained using AdamW\cite{loshchilov2019decoupledweightdecayregularization} optimizer, with a learning rate of $1e^{-3}$.

We use a bilateral grid to color-match the albedo map between the cross-polarized and parallel-polarized streams. The size of the learnable bilateral grid is $16 \times 16\times 8\times12$ which is optimized via AdamW optimizer with a learning rate equal to $1e^{-3}$.

Each sequence is trained for $16k$ iterations in the first stage and $100k$ iterations in the second stage. All our experiments were performed on an Nvidia RTX A6000 GPU.
\section{Experiments \& Results}

\begin{figure*}
    \centering
    \includegraphics[width=0.9\textwidth]{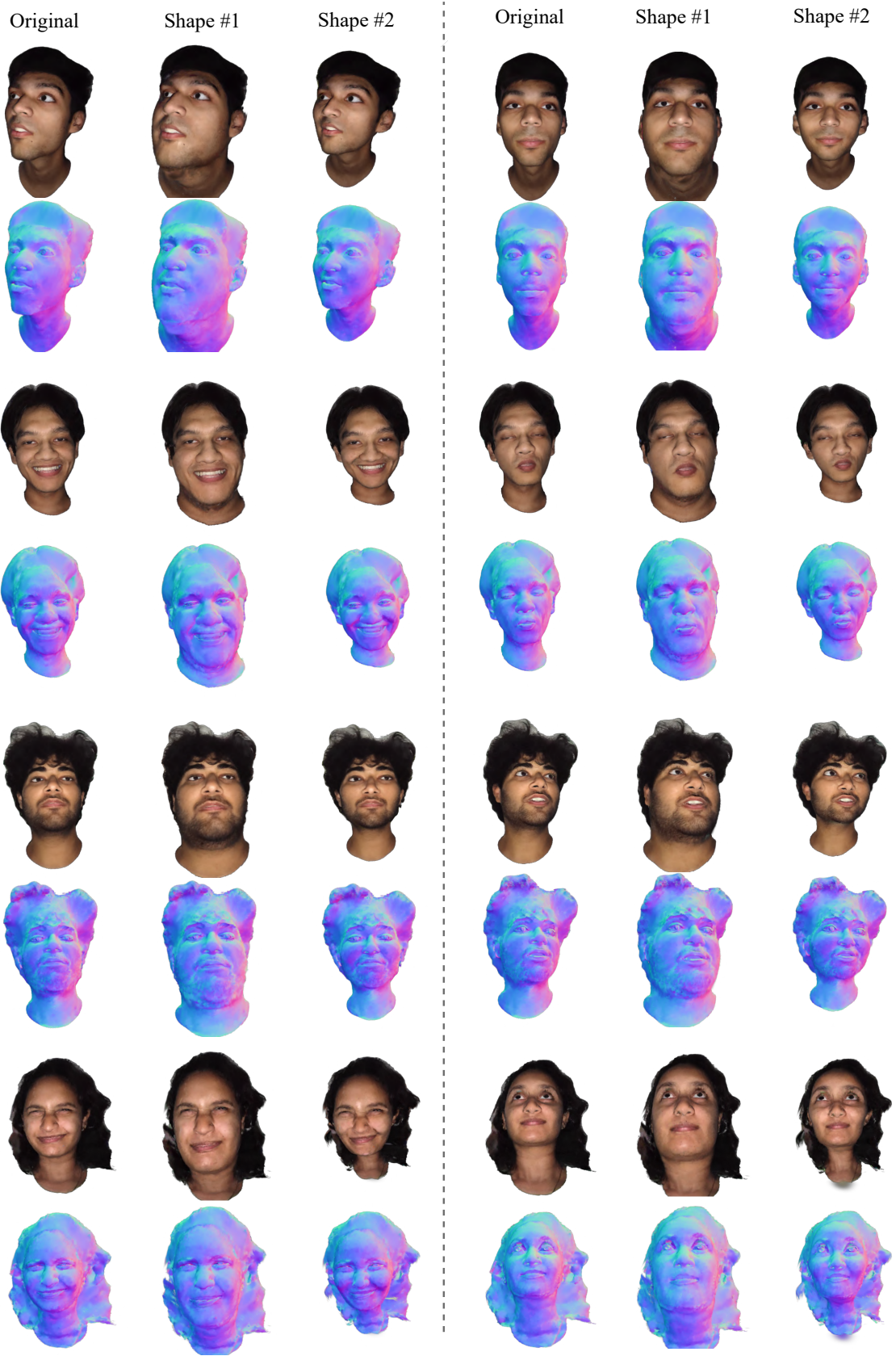}
    \caption{Shape editing over reconstructed head avatars.}
    \label{fig:shape_edit}
\end{figure*}

\begin{table*}[t]
    % \centering
    \centering
    
      \caption{Comparison of different methods on INSTA, 3DGB and Our Dataset. Best values are highlighted in green, while second-best values are in yellow.}
    \centering
    \renewcommand{\arraystretch}{1.2}
    \setlength{\tabcolsep}{4pt}
    \newcommand{\hlfirst}[1]{\cellcolor{green!30} #1}  % Best values
    \newcommand{\hlsecond}[1]{\cellcolor{yellow!50} #1}    % Second best values

    \scriptsize
    \begin{tabular}{l c c c | c c c | c c c}
        \toprule
        & \multicolumn{3}{c}{\textbf{INSTA\cite{mueller2022instant}}} 
        & \multicolumn{3}{c}{\textbf{3DGB \cite{ma2024gaussianblendshapes}}} 
         & \multicolumn{3}{c}{\textbf{DuoPolo (Ours))}}\\
        \cmidrule(lr){2-4} \cmidrule(lr){5-7} \cmidrule(lr){8-10}
        Method & PSNR$\uparrow$ & SSIM$\uparrow$ & LPIPS$\downarrow$ %& Train time$\downarrow$ & Memory$\downarrow$ 
               & PSNR$\uparrow$ & SSIM$\uparrow$ & LPIPS$\downarrow$ %& Train time$\downarrow$ & Memory$\downarrow$ 
               & PSNR$\uparrow$ & SSIM$\uparrow$ & LPIPS$\downarrow$\\ %& Train time$\downarrow$ & Memory$\downarrow$ \\
        \midrule
        Point-Avatar  & 29.12 & 0.932 & 0.094 %& 7.50 m & 48 MB  
                          & 31.76 & 0.926 & 0.145 %& 8.0 m & 48 MB  
                          & 27.77 & 0.919 & 0.128 \\%& 8.0 m & 48 MB  \\
        FlashAvatar & \hlsecond{30.14} & 0.942 & \hlsecond{0.038} %& 26.71 m & 2.1 GB 
                  & 25.92 & 0.920 & 0.094 %& ?? m & 2.7 GB
                   & 29.79 & 0.923 & \hlsecond{0.089}\\ %& 27.82 m & 2.7 GB\\
        Gaussian Deja-vu & 25.56 & 0.925 & 0.058 %& 11.43 m & 640 MB  
             & 23.45 & 0.910 & \hlsecond{0.079} %& ?? m & 580 MB
             & 24.73 & 0.848 & 0.183 \\%& 11.43 m & 580 MB \\
        SPARK & 23.75 & 0.873 & 0.103 %& 15.78 m & 628 MB  
                     & 24.70 & 0.863 & 0.101 %& 13.6 m & 581 MB 
                     & 23.75 & 0.873 & 0.103\\ %& 13.6 m & 581 MB \\
        GaussianBlendshapes & 30.01 & \hlsecond{0.951} & {0.084} %& 26.08 m & 566 MB  
                      & \hlsecond{32.05} & \hlsecond{0.942} & 0.144 %& 8.12 m & 543 MB 
                      & \hlsecond{30.02} & \hlsecond{0.943} & 0.094 \\%& 8.12 m & 543 MB \\
        \textbf{Ours} & \hlfirst{30.33} & \hlfirst{0.952} & \hlfirst{0.036} %& 11.25 m & \hlfirst{117 MB}  
                        & \hlfirst{32.92} & \hlfirst{0.943} & \hlfirst{0.046} %& 18.05 m & \hlfirst{83 MB}
                        & \hlfirst{31.98} & \hlfirst{0.955} & \hlfirst{0.053} \\%& 18.05 m & \hlfirst{83 MB} \\
        \bottomrule
    \end{tabular}
    % }
 %  \caption{Comparison of different methods on INSTA, 3DGB and Our Dataset. Best values are highlighted in \hlfirst{red}, while second best values are in \hlsecond{yellow}.}
    \label{tab:comparison}
\end{table*}%
\noindent
% \textbf{Training \& Implementation Details:}\\
\textbf{Evaluation Datasets \& Metrics:} For experimental evaluation we use sequences from our proposed DuoPolo dataset. For each sequence, we perform a train-test split where we use the first $80$\% of the frames for training and the remaining $20$\% for testing. For quantitative evaluation, we use standard metrics used by existing state-of-the-art methods. Following the evaluation technique of \cite{Zheng2023pointavatar}, we compute per-sequence (or per-subject) $L_1$ error, PSNR\cite{fardo2016formalevaluationpsnrquality} \& SSIM\cite{5596999} between the rendered frames and input (ground truth) frames from the test split.
\begin{figure}[h]
    \centering
    \includegraphics[width=\linewidth]{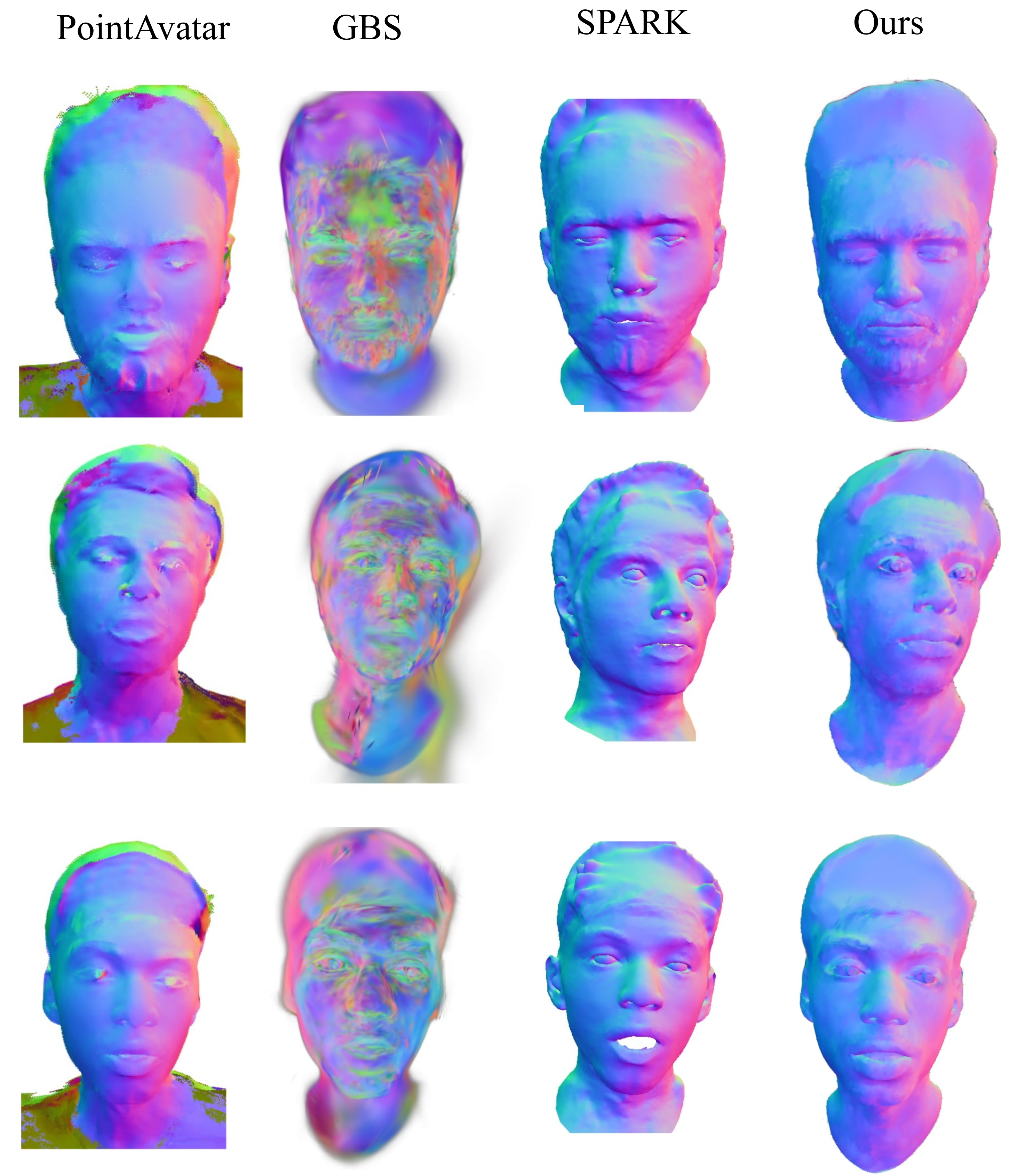}
    \caption{Comparison for surface normals of the reconstructed facial geometry.}
    \label{fig:normal_comparison}
\end{figure}
\begin{figure}[h!]
    \centering
    \includegraphics[width=\linewidth]{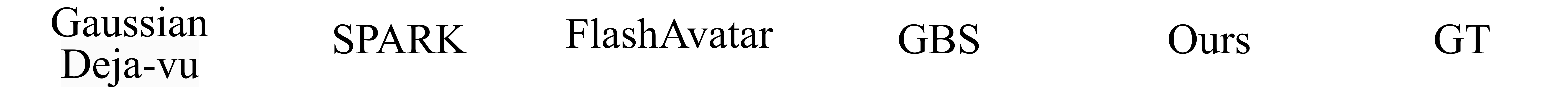}
    \includegraphics[width=\linewidth]{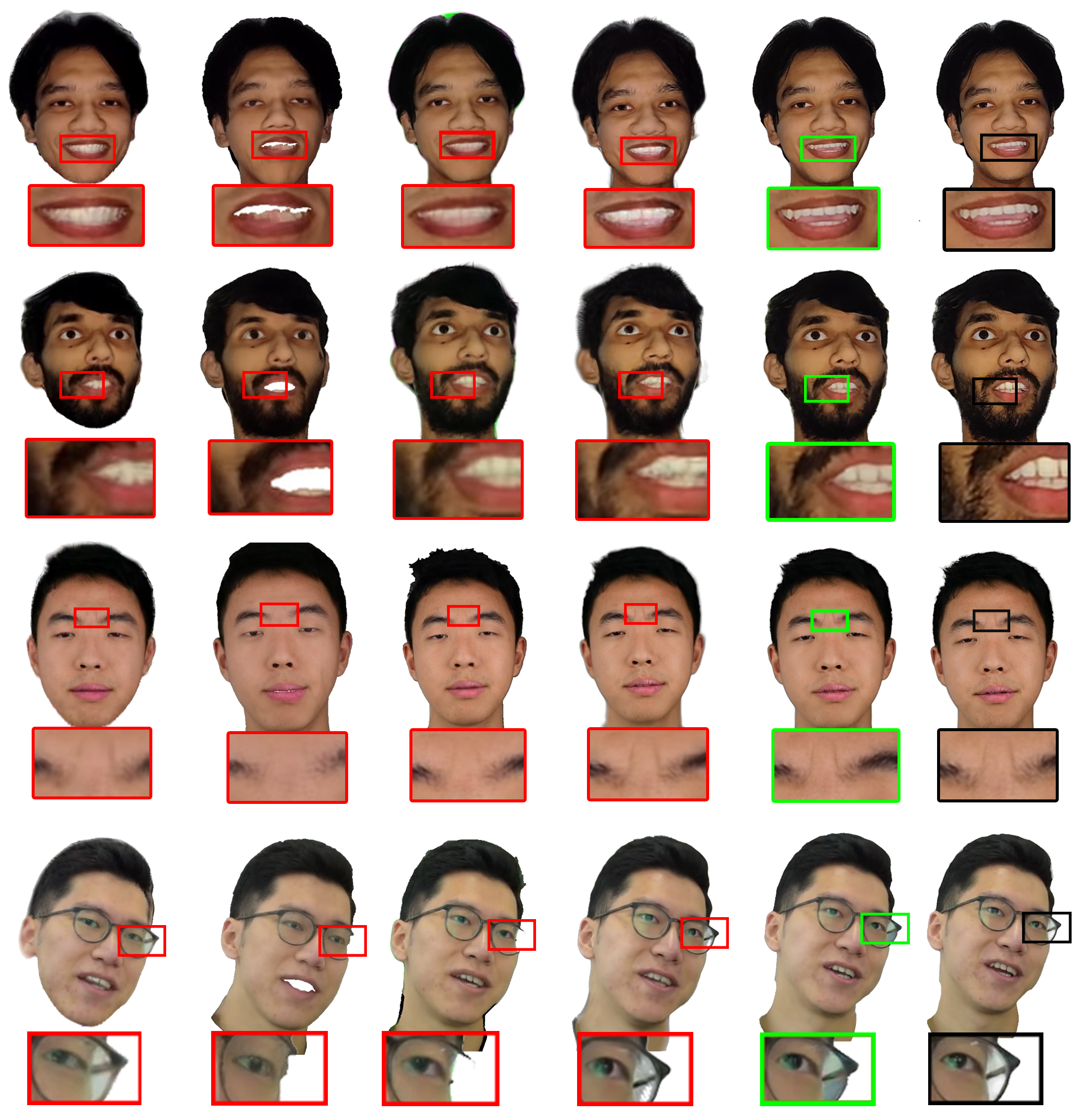}
    \caption{Qualitative comparison with SOTA methods. Our method is able to better capture finer details (such as teeth and eyes) whilst supporting relighting, pose and expression control. }
    \label{fig:comparison_duopolo}
    \vspace{-3mm}
\end{figure}

\subsection{Qualitative Results}
\begin{figure*}[h]
    \centering
    \includegraphics[width=\linewidth, trim={0 0 28cm 0}, clip]{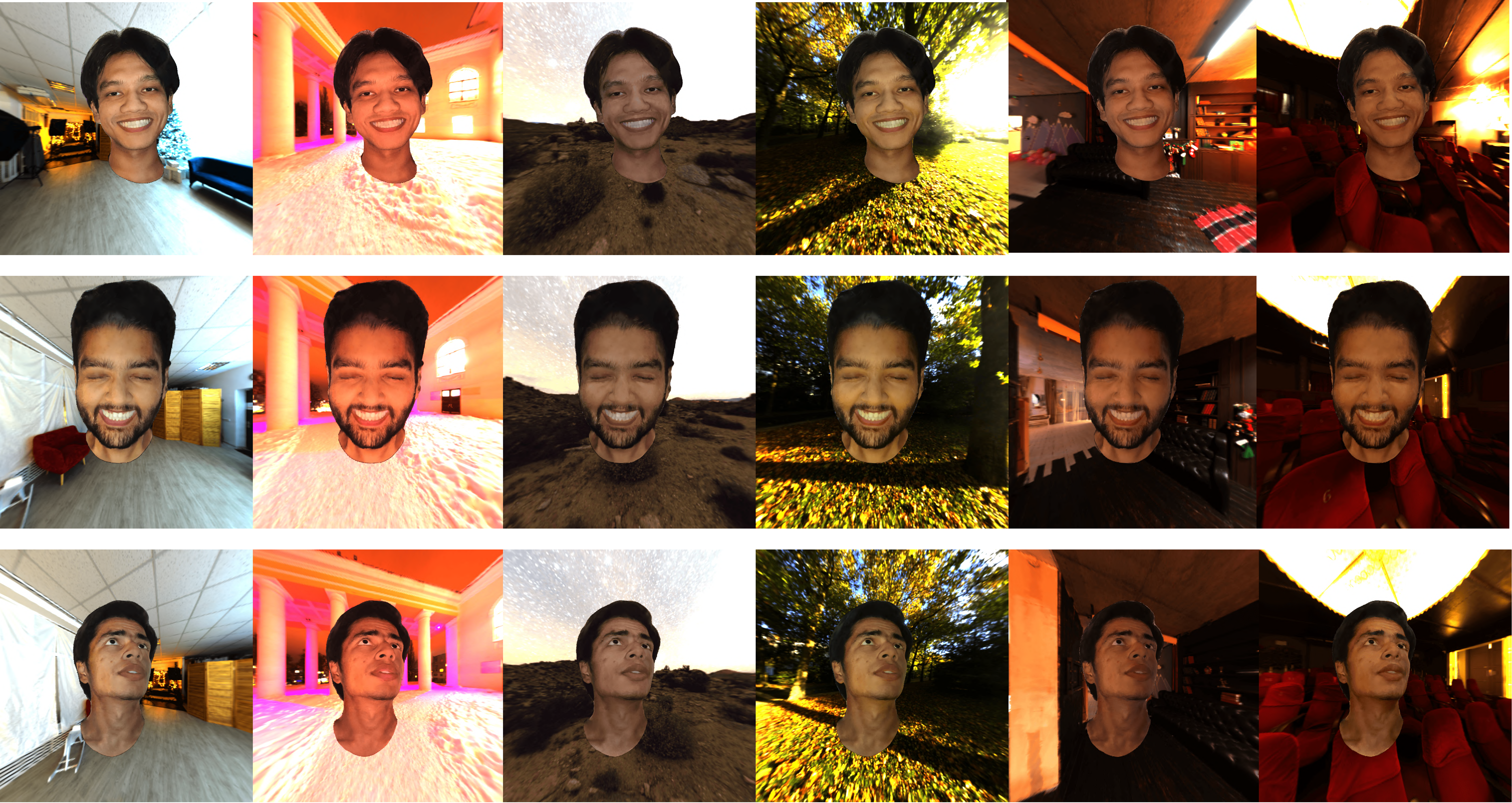}
    \caption{Relighting the reconstructed head avatars in diverse environments.}
    \label{fig:relit}
\end{figure*}

\noindent
\textbf{Learned Reflectance Maps for Relighting:} 
\autoref{fig:qual_ours} demonstrates the ability of the proposed method to capture Physically plausible facial reflectance by separating appearance into albedo, roughness, and normals in the form of UV maps, while also reconstructing high-frequency geometrical details.
% We show the qualitative results of our learning methodlogy in \autoref{fig:qual_ours} on a few sequences captured using the proposed capture setup. As can be seen, the proposed technique has the capability to model the Physically-accurate facial reflectance by disentangling appearance information from the geometry in the form of albedo, roughness and normals $uv$-maps, yielding high-quality renderings while preserving high-frequency geometrical details.\\ 
This disentanglement allows relighting the reconstructed head avatars using diverse environment maps as shown in \autoref{fig:relit}.
\noindent
% \textbf{Relighting:} We show relighting of the learned head avatars in \autoref{fig:ours_relighting}, where we relight the avatars from different light-directions in a novel pose referenced from an original test frame, highlighting the capability of our representation to produce view/pose-consistent relighting.
% 
% 
\\
\noindent
\textbf{Editing:} Our proposed textured human head representation allows shape editing by changing the shape parameters of underlying FLAME mesh (\autoref{fig:shape_edit}), as well as appearance editing, by modifying only the albedo map, similar to classical textured mesh editing. We show an example of appearance editing in \autoref{fig:teaser} where we manipulate the albedo via text using the approach proposed in CLIP-Head \cite{10.1145/3610543.3626169}. 

% appearance editing by directly modifying the albedo texture map,  This approach allows the user to alter the appearance of the head avatar and re-render. We show such application in \autoref{fig:teaser}, where we perform text-based texture editing using via \cite{10.1145/3610543.3626169}, and render the head avatar in novel poses/views.
% 
% such as text-based texture editing models \cite{10.1145/3610543.3626169}. \autoref{fig:teaser} shows such editing performed via inpainting conditioned on text. For more results please refer to the supplementary.
% Please refer to the supplementary additional qualitative/quantitative evaluation. We recommend to see the supplementary video for dynamic results.
% \noindent
% \textbf{Shape Editing}\\
% \noindent
% \textbf{Video}
\subsection{Comparison}
Since most of the learning-based monocular head avatar methods assume a uniformly lit environment with fixed illumination and do not handle learning on polarization data, for fair evaluation, we compare our training methodology with existing SOTA methods on cross-polarized (diffused) sequences from our captured data. In addition to this, we also use samples from INSTA\cite{mueller2022instant} dataset and 3DGB dataset proposed in \cite{ma2024gaussianblendshapes}. We follow the complete evaluation strategy of GaussianBlenshapes (GBS) \cite{ma2024gaussianblendshapes} and report PSNR, LPIPS \& SSIM in \autoref{tab:comparison}, where we outperform all the existing methods for head avatar reconstruction from monocular video. We show qualitative comparison in \autoref{fig:comparison_duopolo} where we show superior quality results compared to others. Our method is able to capture fine details (as highlighted in boxes) while also enabling relighting and editing, which other methods do not directly support.
% We compare with SOTA methods -- PointAvatar\cite{Zheng2023pointavatar} \& GaussianBlendshapes \cite{ma2024gaussianblendshapes} in \autoref{fig:comparison_duopolo} where we outperform both the methods in terms of novel view/pose rendering and detail preservation. Both methods produce blurry and noisy details (highlighted in red) on the novel frames, owing to the multiview inconsistency problem of 3DGS\cite{kerbl3Dgaussians}, while our method uses noise-free high-quality renderings. PointAvatar is also highly prone to tracking noise, resulting in incorrect eyes in \autoref{fig:comparison_duopolo} (c). We also perform quantitative evaluation on the same subjects -- (a), (b) \& (c) and report the results in \autoref{tab:comparison_quant}, where we outperform other methods in majority of the cases.

We also compare the surface normals of the head avatar with some of the methods which aim to reconstruct facial geometry besides appearance in \autoref{fig:normal_comparison}. As shown, PointAvatar produces oversmooth geometry due to inability of point splats to represent detailed geometry. While \cite{Zheng2023pointavatar} GBS\cite{ma2024gaussianblendshapes} produces a lot of spiky artifacts, mainly due to the inherent limitation of 3D Gaussians as they are not suitable for surface reconstruction. SPARK\cite{baert2024spark} models facial geometry as a mesh by learning offsets over the underlying FLAME head mesh, however fails to capture high-frequency details as wrinkles, beard, loose skin etc. On the other hand, our proposed 2DGS-based head representation results in high-quality surface normals capturing subject-specific finer details.
\subsection{Ablation Study}
We perform an ablative analysis of key components proposed in our method. We demonstrate the effect of using residual UV maps $\Delta \Omega$ in \autoref{tab:residual}. We also demonstrate the effect of various choices of low-dimensional projection space for linear basis blend weights (value of $k$) in \autoref{tab:basisk}, where we can observe that chosing higher value for $k$ improves the quality of results, but also significantly increases storage. Increasing the value of $k$ will also increase the number of learning parameters, which will increase the training time as well. We also show the effect of resolution of UV maps in \autoref{tab:tex_res_abl_quant}, where we demonstrate the trade-off between rendering quality and training time. Higher resolution maps require more Gaussians, leading to slow convergence. However, the improvements in rendering quality are not drastic; therefore, we use $512\times512$ as the default resolution.
% 
% \AR{Can we make the table formats consistent?}
% 
% 
% 
% 
\begin{table}[h]
    \centering
    \caption{Effect of Residual Maps}
    \small
    \begin{tabular}{c|c|c|c}
    \hline
    % \textbf{} & \multicolumn{3}{c}{Subject-1} & \multicolumn{3}{c}{Subject-2} \\
    \textbf{Texture Maps} & $\mathbf{PSNR}\uparrow$ & \textbf{SSIM}$\uparrow$ & \textbf{LPIPS}$\downarrow$ \\
    \hline
    w $\Delta\Omega_i$ & $\textbf{33.12}$ & $\textbf{0.953}$ & $\textbf{0.048}$ \\
    % \hline
    w/ o $\Delta\Omega_i$  & $30.99$ & $0.945$ & $0.053$ \\
    \hline
    % Ours w/o $\mathcal{L}_{flame}$ & $0.811$ & $0.948$ & $30.25$\\
    \end{tabular}
    \label{tab:residual}
\end{table}

\begin{table}[h]
    \centering
    \caption{Ablation over values of $k$ for residual basis}
    \resizebox{\columnwidth}{!}{
    \begin{tabular}{c|c|c|c|c}
    \hline
    % \textbf{} & \multicolumn{3}{c}{Subject-1} & \multicolumn{3}{c}{Subject-2} \\
    \textbf{Num. Basis Maps} & $\mathbf{PSNR}$$\uparrow$ & \textbf{SSIM}$\uparrow$ & \textbf{LPIPS}$\downarrow$ & \textbf{Storage(MB)}$\downarrow$\\
    \hline
    \textbf{5} & $32.46$ & $0.953$ & $0.039$ & $\textbf{52.4}$\\
    % \hline
     \textbf{25}  & $\textbf{33.91}$ & $\textbf{0.960}$ & $\textbf{0.035}$ & $262.2$\\
    % \hline
    % Ours w/o $\mathcal{L}_{flame}$ & $0.811$ & $0.948$ & $30.25$\\
    \textbf{50} & $33.62$ & $\textbf{0.960}$ & $0.036$ & $524.3$\\
    % \hline
    \textbf{100} & $33.59$ & $\textbf{0.960}$ & $0.037$ & $1048$\\
    \hline
    \end{tabular}
    }
    \label{tab:basisk}
\end{table}
\begin{table}[h]
    \centering
    \caption{Effect of TexMap resolution }
    \resizebox{\columnwidth}{!}{
    \begin{tabular}{c|c|c|c|c}
    \hline
    % \textbf{} & \multicolumn{3}{c}{Subject-1} & \multicolumn{3}{c}{Subject-2} \\
    \textbf{Texture Resolution} & $\mathbf{PSNR}$$\uparrow$ & \textbf{SSIM}$\uparrow$ & \textbf{LPIPS}$\downarrow$ & \textbf{Time(min)}$\downarrow$\\
    \hline
    128x128 & $33.12$ & $0.953$ & $0.048$ & $5.78$\\
    % \hline
     256x256  & $33.65$ & $0.957$ & $0.039$ & $6.2$\\
    % \hline
    % Ours w/o $\mathcal{L}_{flame}$ & $0.811$ & $0.948$ & $30.25$\\
    \textbf{512x512} & $33.84$ & $0.959$ & $0.036$ & $11.2$\\
    % \hline
    1024x1024 & $33.92$ & $0.961$ & $0.033$ & $28.8$\\
    \hline
    \end{tabular}
    }
    \label{tab:tex_res_abl_quant}
\end{table}
% 
% 
% We perform ablative analysis of several modelling choices incorporated in our training procedure and summarize qualitative results in \autoref{fig:ablation}.\\
% \textbf{Bilateral Grid $\mathcal{U}_\mathrm{W}$:}As demonstrated in the figure, the bilateral grid $\mathcal{U}_\mathrm{W}$ plays a crucial role during second stage by modifying the tint of albedo learned from cross-polarized images to match with tint of  parallel-polarized images.\\
% \noindent
% \textbf{Pose-conditioned MLP $\Phi_{\Delta}$:} As can be seen, $\Phi_{\Delta}$  allows to retain pose-specific fine wrinkles and other details during novel-pose rendering.\\
% \noindent
% \textbf{FLAME-based regularization $\mathcal{L}_{flame}$:} As empirically observed, $\mathcal{L}_{flame}$ in \autoref{eq:loss_function} regularizes the noise in the learned $uv-$maps by forcing the unobserved texels in the initial iterations to match the rendering of FLAME mesh. 
% \section{Discussion}
% \textbf{Limitations}\\ 
% \textbf{Ethical Concerns}
\section{Discussion}
\subsection{Limitations}
Our method has the potential to democratize access to relightable head avatars by making polarized facial performance acquisition more accessible and cost-effective. Using the proposed capture approach, we create a monocular polarized dataset that is unique in its kind. However, its quality cannot be directly compared to data captured using advanced volumetric lightstage systems. The resolution of our dataset is constrained by the limitations of smartphone cameras, which occasionally leads to inaccuracies when modeling fine-grained specular reflections. Additionally, the range of lighting variations is restricted, making it challenging to generalize relighting for complex environment maps.

The proposed representation is further constrained by the parametric head model, which struggles to accurately represent intricate inner-mouth regions such as the tongue and teeth, occasionally resulting in blurry details in those regions. Moreover, the 2DGS representation is unable to capture fine details like individual beard hairs, hair strands, or eyelashes. To address this, we plan to investigate neural strand-based representations for more precise modeling of these facial attributes. Additionally, we aim to explore more accurate techniques for differentiable rendering, such as differentiable ray tracing, as opposed to Gaussian splatting, to better model physics-based reflectance properties.

\subsection{Ethical Concerns}
Our work involves the collection of a dataset, comprising facial information from several individuals. We acknowledge the potential privacy concerns associated with such data and have taken proactive measures to address these issues responsibly. Participation in the dataset collection was entirely voluntary, and we obtained explicit informed consent from all subjects prior to their inclusion. Each participant was fully briefed about the scope, purpose, and potential uses of the dataset, ensuring transparency and understanding. To further safeguard privacy and mitigate risks, we have implemented stringent data access protocols. Access to the dataset is only granted after a thorough review and approval process, requiring researchers to submit a formal application and agree to adhere to ethical guidelines. We emphasize that the dataset will be used solely for academic and research purposes, and its distribution is controlled to prevent misuse.\\
\\
\noindent
While our method has been proposed to advance research in relightable head avatars, we recognize that it could potentially be misused in unethical ways, such as creating deepfake content. Deepfakes, while having legitimate applications in entertainment and education, are frequently associated with harmful consequences, including misinformation, identity theft, and digital harassment. We believe that addressing these ethical considerations is essential to balance the benefits of technological progress with its social implications. By fostering transparency, accountability, and oversight, by restricting the usage of our dataset and approach for research purposes only, we aim to minimize risks while enabling the legitimate and responsible use of our research.
\subsection{Conclusion}
% We present LightHeadEd, a highly scalable and affordable method to capture and reconstruct animatable, relightable and editable head avatars using a commodity smartphone equipped with polaroid filters. We propose a new seamless dynamic acquisition process to capture polaroid facial performances, a novel textured Gaussian head representation with 2D Gaussian attributes embedded within UV space of parametric head mesh, and a novel self-supervised analysis-by-synthesis technique to reconstruct these avatars from monocular video. We demonstrate several useful applications, such as relighting \& text-based appearance editing, while yielding superior quality geometry, appearance and Physically-accurate shading as compared to existing methods. We believe our proposed method aims to bridge the gap between expensive light stage capture system and affordable personalized head avatars accessible to everyone. In future, we wish to adopt our method for higher-resolution images (up to 4K) and to handle finer details such as hair strands, skin pores etc, which we currently don't handle. We also plan to collect a larger dataset across various demographics to build a more generalizable universal prior model.
We proposed LightHeadEd, a scalable and affordable method for capturing and reconstructing animatable, relightable, and editable head avatars using a commodity smartphone with polaroid filters. Our approach introduces a seamless dynamic acquisition process for polaroid facial performances, a textured Gaussian head representation with 2D Gaussian attributes embedded in the UV space of a parametric head mesh, and a self-supervised training scheme to reconstruct head avatars from monocular video. LightHeadEd enables applications like relighting and text-based appearance editing, delivering superior geometry, appearance, and physically-accurate shading compared to existing methods. While finer details like hair strands and skin pores remain current limitations, we plan to address these in the future along with support for higher-resolution images (up to 4K) and the collection of a larger, diverse dataset to facilitate a more generalizable universal head model.
% Limitations: higher resolution in future (as PolFace)
%
% \input{ICCV2025/sec/X_suppl}
% \in
% \clearpage
{
    \small
    \bibliographystyle{unsrtnat}
    \bibliography{main}

\begin{thebibliography}{58}
\providecommand{\natexlab}[1]{#1}
\providecommand{\url}[1]{\texttt{#1}}
\expandafter\ifx\csname urlstyle\endcsname\relax
  \providecommand{\doi}[1]{doi: #1}\else
  \providecommand{\doi}{doi: \begingroup \urlstyle{rm}\Url}\fi

\bibitem[Cao et~al.(2022)Cao, Simon, Kim, Schwartz, Zollhoefer, Saito, Lombardi, Wei, Belko, Yu, Sheikh, and Saragih]{10.1145/3528223.3530143}
Chen Cao, Tomas Simon, Jin~Kyu Kim, Gabe Schwartz, Michael Zollhoefer, Shun-Suke Saito, Stephen Lombardi, Shih-En Wei, Danielle Belko, Shoou-I Yu, Yaser Sheikh, and Jason Saragih.
\newblock Authentic volumetric avatars from a phone scan.
\newblock \emph{ACM Trans. Graph.}, 41\penalty0 (4), July 2022.
\newblock ISSN 0730-0301.
\newblock \doi{10.1145/3528223.3530143}.
\newblock URL \url{https://doi.org/10.1145/3528223.3530143}.

\bibitem[Li et~al.(2024)Li, Cao, Schwartz, Khirodkar, Richardt, Simon, Sheikh, and Saito]{li2024uravatar}
Junxuan Li, Chen Cao, Gabriel Schwartz, Rawal Khirodkar, Christian Richardt, Tomas Simon, Yaser Sheikh, and Shunsuke Saito.
\newblock Uravatar: Universal relightable gaussian codec avatars.
\newblock In \emph{ACM SIGGRAPH 2024 Conference Papers}, 2024.

\bibitem[Baert et~al.(2024)Baert, Bharadwaj, Castan, Maujean, Christie, Abrevaya, and Boukhayma]{baert2024spark}
Kelian Baert, Shrisha Bharadwaj, Fabien Castan, Benoit Maujean, Marc Christie, Victoria Abrevaya, and Adnane Boukhayma.
\newblock Spark: Self-supervised personalized real-time monocular face capture.
\newblock 2024.
\newblock \doi{10.1145/3680528.3687704}.

\bibitem[Zheng et~al.(2023)Zheng, Yifan, Wetzstein, Black, and Hilliges]{Zheng2023pointavatar}
Yufeng Zheng, Wang Yifan, Gordon Wetzstein, Michael~J. Black, and Otmar Hilliges.
\newblock Pointavatar: Deformable point-based head avatars from videos.
\newblock In \emph{Proceedings of the IEEE/CVF Conference on Computer Vision and Pattern Recognition (CVPR)}, 2023.

\bibitem[Gafni et~al.(2021)Gafni, Thies, Zollh{\"o}fer, and Nie{\ss}ner]{Gafni_2021_CVPR}
Guy Gafni, Justus Thies, Michael Zollh{\"o}fer, and Matthias Nie{\ss}ner.
\newblock Dynamic neural radiance fields for monocular 4d facial avatar reconstruction.
\newblock In \emph{Proceedings of the IEEE/CVF Conference on Computer Vision and Pattern Recognition (CVPR)}, pages 8649--8658, June 2021.

\bibitem[Li et~al.(2017)Li, Bolkart, Black, Li, and Romero]{FLAME:SiggraphAsia2017}
Tianye Li, Timo Bolkart, Michael.~J. Black, Hao Li, and Javier Romero.
\newblock Learning a model of facial shape and expression from {4D} scans.
\newblock \emph{ACM Transactions on Graphics, (Proc. SIGGRAPH Asia)}, 36\penalty0 (6):\penalty0 194:1--194:17, 2017.
\newblock URL \url{https://doi.org/10.1145/3130800.3130813}.

\bibitem[Ma et~al.(2024)Ma, Weng, Shao, and Zhou]{ma2024gaussianblendshapes}
Shengjie Ma, Yanlin Weng, Tianjia Shao, and Kun Zhou.
\newblock 3d gaussian blendshapes for head avatar animation.
\newblock In \emph{ACM SIGGRAPH Conference Proceedings, Denver, CO, United States, July 28 - August 1, 2024}, 2024.

\bibitem[Chen et~al.(2023{\natexlab{a}})Chen, Wang, Li, Xiao, Zhang, Yao, and Liu]{chen2023monogaussianavatar}
Yufan Chen, Lizhen Wang, Qijing Li, Hongjiang Xiao, Shengping Zhang, Hongxun Yao, and Yebin Liu.
\newblock Monogaussianavatar: Monocular gaussian point-based head avatar.
\newblock \emph{arXiv}, 2023{\natexlab{a}}.

\bibitem[Kerbl et~al.(2023)Kerbl, Kopanas, Leimk{\"u}hler, and Drettakis]{kerbl3Dgaussians}
Bernhard Kerbl, Georgios Kopanas, Thomas Leimk{\"u}hler, and George Drettakis.
\newblock 3d gaussian splatting for real-time radiance field rendering.
\newblock \emph{ACM Transactions on Graphics}, 42\penalty0 (4), July 2023.
\newblock URL \url{https://repo-sam.inria.fr/fungraph/3d-gaussian-splatting/}.

\bibitem[Huang et~al.(2024)Huang, Yu, Chen, Geiger, and Gao]{Huang2DGS2024}
Binbin Huang, Zehao Yu, Anpei Chen, Andreas Geiger, and Shenghua Gao.
\newblock 2d gaussian splatting for geometrically accurate radiance fields.
\newblock In \emph{SIGGRAPH 2024 Conference Papers}. Association for Computing Machinery, 2024.
\newblock \doi{10.1145/3641519.3657428}.

\bibitem[M{\"u}ller(1995)]{Mller1995PolarizationBasedSO}
Volker M{\"u}ller.
\newblock Polarization-based separation of diffuse and specular surface-reflection.
\newblock In \emph{DAGM-Symposium}, 1995.
\newblock URL \url{https://api.semanticscholar.org/CorpusID:51804148}.

\bibitem[Rahmann and Canterakis(2001)]{990468}
S.~Rahmann and N.~Canterakis.
\newblock Reconstruction of specular surfaces using polarization imaging.
\newblock In \emph{Proceedings of the 2001 IEEE Computer Society Conference on Computer Vision and Pattern Recognition. CVPR 2001}, volume~1, pages I--I, 2001.
\newblock \doi{10.1109/CVPR.2001.990468}.

\bibitem[Wolff and Boult(1991)]{85655}
L.B. Wolff and T.E. Boult.
\newblock Constraining object features using a polarization reflectance model.
\newblock \emph{IEEE Transactions on Pattern Analysis and Machine Intelligence}, 13\penalty0 (7):\penalty0 635--657, 1991.
\newblock \doi{10.1109/34.85655}.

\bibitem[Debevec et~al.(2000)Debevec, Hawkins, Tchou, Duiker, Sarokin, and Sagar]{10.1145/344779.344855}
Paul Debevec, Tim Hawkins, Chris Tchou, Haarm-Pieter Duiker, Westley Sarokin, and Mark Sagar.
\newblock Acquiring the reflectance field of a human face.
\newblock In \emph{Proceedings of the 27th Annual Conference on Computer Graphics and Interactive Techniques}, SIGGRAPH '00, page 145–156, USA, 2000. ACM Press/Addison-Wesley Publishing Co.
\newblock ISBN 1581132085.
\newblock \doi{10.1145/344779.344855}.
\newblock URL \url{https://doi.org/10.1145/344779.344855}.

\bibitem[Ghosh et~al.(2011)Ghosh, Fyffe, Tunwattanapong, Busch, Yu, and Debevec]{10.1145/2070781.2024163}
Abhijeet Ghosh, Graham Fyffe, Borom Tunwattanapong, Jay Busch, Xueming Yu, and Paul Debevec.
\newblock Multiview face capture using polarized spherical gradient illumination.
\newblock \emph{ACM Trans. Graph.}, 30\penalty0 (6):\penalty0 1–10, December 2011.
\newblock ISSN 0730-0301.
\newblock \doi{10.1145/2070781.2024163}.
\newblock URL \url{https://doi.org/10.1145/2070781.2024163}.

\bibitem[Wilson et~al.(2010)Wilson, Ghosh, Peers, Chiang, Busch, and Debevec]{10.1145/1731047.1731055}
Cyrus~A. Wilson, Abhijeet Ghosh, Pieter Peers, Jen-Yuan Chiang, Jay Busch, and Paul Debevec.
\newblock Temporal upsampling of performance geometry using photometric alignment.
\newblock \emph{ACM Trans. Graph.}, 29\penalty0 (2), April 2010.
\newblock ISSN 0730-0301.
\newblock \doi{10.1145/1731047.1731055}.
\newblock URL \url{https://doi.org/10.1145/1731047.1731055}.

\bibitem[Woodham(1980)]{Woodham1980PhotometricMF}
Robert~J. Woodham.
\newblock Photometric method for determining surface orientation from multiple images.
\newblock 1980.
\newblock URL \url{https://api.semanticscholar.org/CorpusID:61691075}.

\bibitem[Gotardo et~al.(2018)Gotardo, Riviere, Bradley, Ghosh, and Beeler]{10.1145/3272127.3275073}
Paulo Gotardo, J\'{e}r\'{e}my Riviere, Derek Bradley, Abhijeet Ghosh, and Thabo Beeler.
\newblock Practical dynamic facial appearance modeling and acquisition.
\newblock 37\penalty0 (6), December 2018.
\newblock ISSN 0730-0301.
\newblock \doi{10.1145/3272127.3275073}.
\newblock URL \url{https://doi.org/10.1145/3272127.3275073}.

\bibitem[Riviere et~al.(2020)Riviere, Gotardo, Bradley, Ghosh, and Beeler]{10.1145/3386569.3392464}
J\'{e}r\'{e}my Riviere, Paulo Gotardo, Derek Bradley, Abhijeet Ghosh, and Thabo Beeler.
\newblock Single-shot high-quality facial geometry and skin appearance capture.
\newblock \emph{ACM Trans. Graph.}, 39\penalty0 (4), August 2020.
\newblock ISSN 0730-0301.
\newblock \doi{10.1145/3386569.3392464}.
\newblock URL \url{https://doi.org/10.1145/3386569.3392464}.

\bibitem[Xu et~al.(2022)Xu, Riviere, Zoss, Chandran, Bradley, and Gotardo]{10.2312:egs.20221019}
Yingyan Xu, Jérémy Riviere, Gaspard Zoss, Prashanth Chandran, Derek Bradley, and Paulo Gotardo.
\newblock {Improved Lighting Models for Facial Appearance Capture}.
\newblock In Nuria Pelechano and David Vanderhaeghe, editors, \emph{Eurographics 2022 - Short Papers}. The Eurographics Association, 2022.
\newblock ISBN 978-3-03868-169-4.
\newblock \doi{10.2312/egs.20221019}.

\bibitem[Blanz and Vetter(1999)]{10.1145/311535.311556}
Volker Blanz and Thomas Vetter.
\newblock A morphable model for the synthesis of 3d faces.
\newblock In \emph{Proceedings of the 26th Annual Conference on Computer Graphics and Interactive Techniques}, SIGGRAPH '99, page 187–194, USA, 1999. ACM Press/Addison-Wesley Publishing Co.
\newblock ISBN 0201485605.
\newblock \doi{10.1145/311535.311556}.
\newblock URL \url{https://doi.org/10.1145/311535.311556}.

\bibitem[Pighin et~al.(1998)Pighin, Hecker, Lischinski, Szeliski, and Salesin]{Pighin1998SynthesizingRF}
Fr{\'e}d{\'e}ric~H. Pighin, Jamie Hecker, Dani Lischinski, Richard Szeliski, and D.~Salesin.
\newblock Synthesizing realistic facial expressions from photographs.
\newblock \emph{Proceedings of the 25th annual conference on Computer graphics and interactive techniques}, 1998.
\newblock URL \url{https://api.semanticscholar.org/CorpusID:74926}.

\bibitem[Cao et~al.(2015)Cao, Bradley, Zhou, and Beeler]{Cao2015RealtimeHF}
Chen Cao, Derek Bradley, Kun Zhou, and Thabo Beeler.
\newblock Real-time high-fidelity facial performance capture.
\newblock \emph{ACM Transactions on Graphics (TOG)}, 34:\penalty0 1 -- 9, 2015.
\newblock URL \url{https://api.semanticscholar.org/CorpusID:207226489}.

\bibitem[Vlasic et~al.(2004)Vlasic, Brand, Pfister, and Popović]{multilinearvlasic}
Daniel Vlasic, Matthew Brand, Hanspeter Pfister, and Jovan Popović.
\newblock Multilinear models for face synthesis.
\newblock 01 2004.
\newblock \doi{10.1145/1186223.1186293}.

\bibitem[Ranjan et~al.(2018)Ranjan, Bolkart, Sanyal, and Black]{DBLP:journals/corr/abs-1807-10267}
Anurag Ranjan, Timo Bolkart, Soubhik Sanyal, and Michael~J. Black.
\newblock Generating 3d faces using convolutional mesh autoencoders.
\newblock \emph{CoRR}, abs/1807.10267, 2018.
\newblock URL \url{http://arxiv.org/abs/1807.10267}.

\bibitem[Tran and Liu(2018)]{DBLP:journals/corr/abs-1804-03786}
Luan Tran and Xiaoming Liu.
\newblock Nonlinear 3d face morphable model.
\newblock \emph{CoRR}, abs/1804.03786, 2018.
\newblock URL \url{http://arxiv.org/abs/1804.03786}.

\bibitem[Kirschstein et~al.(2023{\natexlab{a}})Kirschstein, Qian, Giebenhain, Walter, and Nie\ss{}ner]{kirschstein2023nersemble}
Tobias Kirschstein, Shenhan Qian, Simon Giebenhain, Tim Walter, and Matthias Nie\ss{}ner.
\newblock Nersemble: Multi-view radiance field reconstruction of human heads.
\newblock \emph{ACM Trans. Graph.}, 42\penalty0 (4), jul 2023{\natexlab{a}}.
\newblock ISSN 0730-0301.
\newblock \doi{10.1145/3592455}.
\newblock URL \url{https://doi.org/10.1145/3592455}.

\bibitem[Zhao et~al.(2023)Zhao, Wang, Sun, Zhang, Suo, and Liu]{zhao2023havatar}
Xiaochen Zhao, Lizhen Wang, Jingxiang Sun, Hongwen Zhang, Jinli Suo, and Yebin Liu.
\newblock Havatar: High-fidelity head avatar via facial model conditioned neural radiance field.
\newblock \emph{ACM Trans. Graph.}, oct 2023.
\newblock ISSN 0730-0301.
\newblock \doi{10.1145/3626316}.
\newblock URL \url{https://doi.org/10.1145/3626316}.
\newblock Just Accepted.

\bibitem[Kirschstein et~al.(2023{\natexlab{b}})Kirschstein, Giebenhain, and Nie{\ss}ner]{kirschstein2023diffusionavatars}
Tobias Kirschstein, Simon Giebenhain, and Matthias Nie{\ss}ner.
\newblock Diffusionavatars: Deferred diffusion for high-fidelity 3d head avatars.
\newblock \emph{arXiv preprint arXiv:2311.18635}, 2023{\natexlab{b}}.

\bibitem[Qian et~al.(2023)Qian, Kirschstein, Schoneveld, Davoli, Giebenhain, and Nie\ss{}ner]{qian2023gaussianavatars}
Shenhan Qian, Tobias Kirschstein, Liam Schoneveld, Davide Davoli, Simon Giebenhain, and Matthias Nie\ss{}ner.
\newblock Gaussianavatars: Photorealistic head avatars with rigged 3d gaussians.
\newblock \emph{arXiv preprint arXiv:2312.02069}, 2023.

\bibitem[Giebenhain et~al.(2024{\natexlab{a}})Giebenhain, Kirschstein, R{\"{u}}nz, Agapito, and Nie{\ss}ner]{giebenhain2024npga}
Simon Giebenhain, Tobias Kirschstein, Martin R{\"{u}}nz, Lourdes Agapito, and Matthias Nie{\ss}ner.
\newblock Npga: Neural parametric gaussian avatars.
\newblock In \emph{SIGGRAPH Asia 2024 Conference Papers (SA Conference Papers '24), December 3-6, Tokyo, Japan}, 2024{\natexlab{a}}.
\newblock ISBN 979-8-4007-1131-2/24/12.
\newblock \doi{10.1145/3680528.3687689}.

\bibitem[Xu et~al.(2024)Xu, Chen, Li, Zhang, Wang, Zheng, and Liu]{xu2023gaussianheadavatar}
Yuelang Xu, Benwang Chen, Zhe Li, Hongwen Zhang, Lizhen Wang, Zerong Zheng, and Yebin Liu.
\newblock Gaussian head avatar: Ultra high-fidelity head avatar via dynamic gaussians.
\newblock In \emph{Proceedings of the IEEE/CVF Conference on Computer Vision and Pattern Recognition (CVPR)}, 2024.

\bibitem[Saito et~al.(2024)Saito, Schwartz, Simon, Li, and Nam]{saito2024rgca}
Shunsuke Saito, Gabriel Schwartz, Tomas Simon, Junxuan Li, and Giljoo Nam.
\newblock Relightable gaussian codec avatars.
\newblock In \emph{CVPR}, 2024.

\bibitem[Manu et~al.(2023)Manu, Srivastava, and Sharma]{10.1145/3610543.3626169}
Pranav Manu, Astitva Srivastava, and Avinash Sharma.
\newblock Clip-head: Text-guided generation of textured neural parametric 3d head models.
\newblock In \emph{SIGGRAPH Asia 2023 Technical Communications}, SA '23, New York, NY, USA, 2023. Association for Computing Machinery.
\newblock ISBN 9798400703140.
\newblock \doi{10.1145/3610543.3626169}.
\newblock URL \url{https://doi.org/10.1145/3610543.3626169}.

\bibitem[Srivastava et~al.(2024)Srivastava, Manu, Raj, Jampani, and Sharma]{srivastava2024wordrobetextguidedgenerationtextured}
Astitva Srivastava, Pranav Manu, Amit Raj, Varun Jampani, and Avinash Sharma.
\newblock Wordrobe: Text-guided generation of textured 3d garments, 2024.
\newblock URL \url{https://arxiv.org/abs/2403.17541}.

\bibitem[Chen et~al.(2023{\natexlab{b}})Chen, Siddiqui, Lee, Tulyakov, and Nie{\ss}ner]{chen2023text2tex}
Dave~Zhenyu Chen, Yawar Siddiqui, Hsin-Ying Lee, Sergey Tulyakov, and Matthias Nie{\ss}ner.
\newblock Text2tex: Text-driven texture synthesis via diffusion models.
\newblock \emph{arXiv preprint arXiv:2303.11396}, 2023{\natexlab{b}}.

\bibitem[Haque et~al.(2023)Haque, Tancik, Efros, Holynski, and Kanazawa]{haque2023instructnerf2nerfediting3dscenes}
Ayaan Haque, Matthew Tancik, Alexei~A. Efros, Aleksander Holynski, and Angjoo Kanazawa.
\newblock Instruct-nerf2nerf: Editing 3d scenes with instructions, 2023.
\newblock URL \url{https://arxiv.org/abs/2303.12789}.

\bibitem[Mendiratta et~al.(2023)Mendiratta, Pan, Elgharib, Teotia, R, Tewari, Golyanik, Kortylewski, and Theobalt]{mendiratta2023avatarstudiotextdrivenediting3d}
Mohit Mendiratta, Xingang Pan, Mohamed Elgharib, Kartik Teotia, Mallikarjun~B R, Ayush Tewari, Vladislav Golyanik, Adam Kortylewski, and Christian Theobalt.
\newblock Avatarstudio: Text-driven editing of 3d dynamic human head avatars, 2023.
\newblock URL \url{https://arxiv.org/abs/2306.00547}.

\bibitem[Wang et~al.(2024{\natexlab{a}})Wang, Kang, Sun, Qian, Wang, Bao, and Zhang]{wang2024mega}
Cong Wang, Di~Kang, He-Yi Sun, Shen-Han Qian, Zi-Xuan Wang, Linchao Bao, and Song-Hai Zhang.
\newblock Mega: Hybrid mesh-gaussian head avatar for high-fidelity rendering and head editing.
\newblock \emph{arXiv preprint arXiv:2404.19026}, 2024{\natexlab{a}}.

\bibitem[Giebenhain et~al.(2024{\natexlab{b}})Giebenhain, Kirschstein, Georgopoulos, R{\"{u}}nz, Agapito, and Nie{\ss}ner]{giebenhain2024mononphm}
Simon Giebenhain, Tobias Kirschstein, Markos Georgopoulos, Martin R{\"{u}}nz, Lourdes Agapito, and Matthias Nie{\ss}ner.
\newblock Mononphm: Dynamic head reconstruction from monoculuar videos.
\newblock In \emph{Proc. IEEE Conf. on Computer Vision and Pattern Recognition (CVPR)}, 2024{\natexlab{b}}.

\bibitem[Shao et~al.(2024)Shao, Wang, Li, Wang, Lin, Zhang, Fan, and Wang]{shao2024splattingavatar}
Zhijing Shao, Zhaolong Wang, Zhuang Li, Duotun Wang, Xiangru Lin, Yu~Zhang, Mingming Fan, and Zeyu Wang.
\newblock {SplattingAvatar: Realistic Real-Time Human Avatars with Mesh-Embedded Gaussian Splatting}.
\newblock In \emph{Proceedings of the IEEE/CVF Conference on Computer Vision and Pattern Recognition (CVPR)}, 2024.

\bibitem[Xiang et~al.(2024)Xiang, Gao, Guo, and Zhang]{xiang2024flashavatarhighfidelityheadavatar}
Jun Xiang, Xuan Gao, Yudong Guo, and Juyong Zhang.
\newblock Flashavatar: High-fidelity head avatar with efficient gaussian embedding, 2024.
\newblock URL \url{https://arxiv.org/abs/2312.02214}.

\bibitem[Azinovi\'c et~al.(2023)Azinovi\'c, Maury, Hery, Nie{\ss}ner, and Thies]{azinovic2022polface}
Dejan Azinovi\'c, Olivier Maury, Christophe Hery, Matthias Nie{\ss}ner, and Justus Thies.
\newblock High-res facial appearance capture from polarized smartphone images.
\newblock In \emph{Proceedings of the IEEE/CVF Conference on Computer Vision and Pattern Recognition (CVPR)}, June 2023.

\bibitem[Qian(2024)]{qian2024versatile}
Shenhan Qian.
\newblock Versatile head alignment with adaptive appearance priors.
\newblock September 2024.
\newblock URL \url{https://github.com/ShenhanQian/VHAP}.

\bibitem[Qian et~al.(2024)Qian, Kirschstein, Schoneveld, Davoli, Giebenhain, and Nie{\ss}ner]{qian2024gaussianavatars}
Shenhan Qian, Tobias Kirschstein, Liam Schoneveld, Davide Davoli, Simon Giebenhain, and Matthias Nie{\ss}ner.
\newblock Gaussianavatars: Photorealistic head avatars with rigged 3d gaussians.
\newblock In \emph{Proceedings of the IEEE/CVF Conference on Computer Vision and Pattern Recognition}, pages 20299--20309, 2024.

\bibitem[Fridovich-Keil et~al.(2022)Fridovich-Keil, Yu, Tancik, Chen, Recht, and Kanazawa]{Fridovich-Keil_2022_CVPR}
Sara Fridovich-Keil, Alex Yu, Matthew Tancik, Qinhong Chen, Benjamin Recht, and Angjoo Kanazawa.
\newblock Plenoxels: Radiance fields without neural networks.
\newblock In \emph{Proceedings of the IEEE/CVF Conference on Computer Vision and Pattern Recognition (CVPR)}, pages 5501--5510, June 2022.

\bibitem[{Nicodemus}(1965)]{1965ApOpt...4..767N}
Fred~E. {Nicodemus}.
\newblock {Directional reflectance and emissivity of an opaque surface}.
\newblock \emph{Applied Optics}, 4\penalty0 (7):\penalty0 767, July 1965.
\newblock \doi{10.1364/AO.4.000767}.

\bibitem[Cook and Torrance(1982)]{10.1145/357290.357293}
R.~L. Cook and K.~E. Torrance.
\newblock A reflectance model for computer graphics.
\newblock \emph{ACM Trans. Graph.}, 1\penalty0 (1):\penalty0 7–24, January 1982.
\newblock ISSN 0730-0301.
\newblock \doi{10.1145/357290.357293}.
\newblock URL \url{https://doi.org/10.1145/357290.357293}.

\bibitem[Schlick(1994)]{https://doi.org/10.1111/1467-8659.1330233}
Christophe Schlick.
\newblock An inexpensive brdf model for physically-based rendering.
\newblock \emph{Computer Graphics Forum}, 13\penalty0 (3):\penalty0 233--246, 1994.
\newblock \doi{https://doi.org/10.1111/1467-8659.1330233}.
\newblock URL \url{https://onlinelibrary.wiley.com/doi/abs/10.1111/1467-8659.1330233}.

\bibitem[Walter et~al.(2007)Walter, Marschner, Li, and Torrance]{10.5555/2383847.2383874}
Bruce Walter, Stephen~R. Marschner, Hongsong Li, and Kenneth~E. Torrance.
\newblock Microfacet models for refraction through rough surfaces.
\newblock In \emph{Proceedings of the 18th Eurographics Conference on Rendering Techniques}, EGSR'07, page 195–206, Goslar, DEU, 2007. Eurographics Association.
\newblock ISBN 9783905673524.

\bibitem[Smith(1967)]{smith-ggx}
B.~Smith.
\newblock Geometrical shadowing of a random rough surface.
\newblock \emph{IEEE Transactions on Antennas and Propagation}, 15\penalty0 (5):\penalty0 668--671, 1967.
\newblock \doi{10.1109/TAP.1967.1138991}.

\bibitem[Ashikhmin and Shirley(2001)]{shirley}
Michael Ashikhmin and Peter Shirley.
\newblock An anisotropic phong light reflection model.
\newblock \emph{Journal of Graphics Tools}, 5, 01 2001.

\bibitem[Munkberg et~al.(2022)Munkberg, Hasselgren, Shen, Gao, Chen, Evans, M\"uller, and Fidler]{Munkberg_2022_CVPR}
Jacob Munkberg, Jon Hasselgren, Tianchang Shen, Jun Gao, Wenzheng Chen, Alex Evans, Thomas M\"uller, and Sanja Fidler.
\newblock {Extracting Triangular 3D Models, Materials, and Lighting From Images}.
\newblock In \emph{Proceedings of the IEEE/CVF Conference on Computer Vision and Pattern Recognition (CVPR)}, pages 8280--8290, June 2022.

\bibitem[M\"uller et~al.(2022)M\"uller, Evans, Schied, and Keller]{mueller2022instant}
Thomas M\"uller, Alex Evans, Christoph Schied, and Alexander Keller.
\newblock Instant neural graphics primitives with a multiresolution hash encoding.
\newblock \emph{ACM Trans. Graph.}, 41\penalty0 (4):\penalty0 102:1--102:15, July 2022.
\newblock \doi{10.1145/3528223.3530127}.
\newblock URL \url{https://doi.org/10.1145/3528223.3530127}.

\bibitem[Wang et~al.(2024{\natexlab{b}})Wang, Wang, Gong, and Xue]{wang2024bilateralguidedradiancefield}
Yuehao Wang, Chaoyi Wang, Bingchen Gong, and Tianfan Xue.
\newblock Bilateral guided radiance field processing, 2024{\natexlab{b}}.
\newblock URL \url{https://arxiv.org/abs/2406.00448}.

\bibitem[Loshchilov and Hutter(2019)]{loshchilov2019decoupledweightdecayregularization}
Ilya Loshchilov and Frank Hutter.
\newblock Decoupled weight decay regularization, 2019.
\newblock URL \url{https://arxiv.org/abs/1711.05101}.

\bibitem[Fardo et~al.(2016)Fardo, Conforto, de~Oliveira, and Rodrigues]{fardo2016formalevaluationpsnrquality}
Fernando~A. Fardo, Victor~H. Conforto, Francisco~C. de~Oliveira, and Paulo~S. Rodrigues.
\newblock A formal evaluation of psnr as quality measurement parameter for image segmentation algorithms, 2016.
\newblock URL \url{https://arxiv.org/abs/1605.07116}.

\bibitem[Horé and Ziou(2010)]{5596999}
Alain Horé and Djemel Ziou.
\newblock Image quality metrics: Psnr vs. ssim.
\newblock In \emph{2010 20th International Conference on Pattern Recognition}, pages 2366--2369, 2010.
\newblock \doi{10.1109/ICPR.2010.579}.

\end{thebibliography}
}

\end{document}